\documentclass{article}

\usepackage{microtype}
\usepackage{graphicx}
\usepackage{subfigure}
\usepackage{booktabs} 

\usepackage{amsmath, amssymb}
\usepackage{algorithm}
\usepackage{algorithmic}
\usepackage{graphicx}
\usepackage{multirow}
\usepackage{color}
\usepackage{wrapfig}
\usepackage{lipsum}
\usepackage{listings}
\usepackage{paralist}
\usepackage{wrapfig}
\usepackage{amsthm}
\usepackage{booktabs}

\usepackage{hyperref}

\newtheorem{theorem}{Theorem}

\newtheorem{lemma}{Lemma}

\newcommand{\fig}[1]{Figure~#1}
\newcommand{\alg}[1]{Algorithm~#1}

\newcommand{\tab}[1]{Table~#1}

\newcommand{\reals}{\mathbb{R}}
\newcommand{\M}{\mathcal{M}}

\usepackage[accepted]{icml2021}

\icmltitlerunning{Density Constrained Reinforcement Learning}
\begin{document}

\twocolumn[
\icmltitle{Density Constrained Reinforcement Learning}



\icmlsetsymbol{equal}{*}

\begin{icmlauthorlist}
\icmlauthor{Zengyi Qin}{MIT}
\icmlauthor{Yuxiao Chen}{Caltech}
\icmlauthor{Chuchu Fan}{MIT}
\end{icmlauthorlist}

\icmlaffiliation{MIT}{Massachusetts Institute of Technology}
\icmlaffiliation{Caltech}{California Institute of Technology}

\icmlcorrespondingauthor{Chuchu Fan}{chuchu@mit.edu}

\icmlkeywords{Machine Learning, ICML}
\vskip 0.3in

]



\printAffiliationsAndNotice{}  

\begin{abstract}

We study constrained reinforcement learning (CRL) from a novel perspective by setting constraints directly on \emph{state density functions}, rather than the value functions considered by previous works. State density has a clear physical and mathematical interpretation, and is able to express a wide variety of constraints such as resource limits and safety requirements. Density constraints can also avoid the time-consuming process of designing and tuning cost functions required by value function-based constraints to encode system specifications. We leverage the duality between density functions and Q functions to develop an effective algorithm to solve the density constrained RL problem optimally and the constrains are guaranteed to be satisfied. We prove that the proposed algorithm converges to a near-optimal solution with a bounded error even when the policy update is imperfect. We use a set of comprehensive experiments to demonstrate the advantages of our approach over state-of-the-art CRL methods, with a wide range of density constrained tasks as well as standard CRL benchmarks such as Safety-Gym.

\end{abstract}
\section{Introduction}

Constrained reinforcement learning (CRL)~\citep{altman1999constrained, achiam2017cpo, dalal2018safe, paternain2019constrained, tessler2019reward, Yang2020Projection, zhang2020first, stooke2020responsive} aims to find the optimal policy that maximizes the cumulative reward signal while respecting certain constraints such as safety requirements and resource limits. Existing CRL approaches typically involve constructing suitable cost functions and value functions to take into account the constraints. Then a crucial step is to choose appropriate parameters such as thresholds for the cost and value functions to encode the constraints. However, one significant gap between the use of such methods and solving CRL problems is the correct construction of the cost and value functions, which is typically not solved systematically but relies on engineering intuitions~\citep{paternain2019constrained}. Simple cost functions may not exhibit satisfactory performance, while sophisticated cost functions may not have clear physical meanings. When cost functions lack clear physical interpretations, it is difficult to formally guarantee the satisfaction of the performance specifications, even if the constraints on the cost functions are fulfilled. Moreover, different environments generally need different cost functions, which makes the tuning process time-consuming.

\begin{figure}
    \centering
    \includegraphics[width=0.48\textwidth]{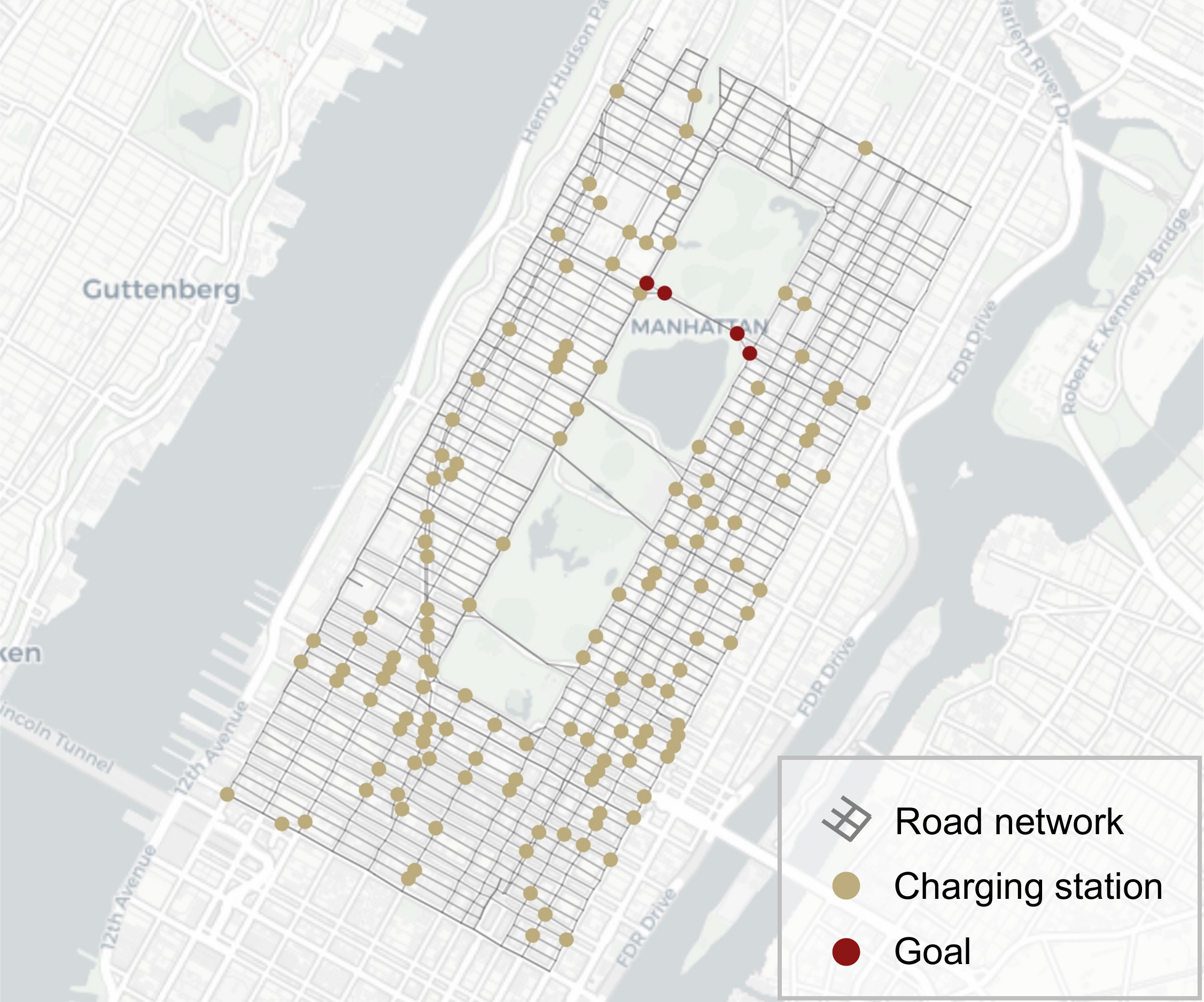}
    \caption{\footnotesize Autonomous electric vehicle routing in Manhattan: Control the electric vehicles to reach the goals and always keep the remaining energy above a threshold. Vehicles can be charged at the charging stations. The \emph{vehicle density} at each charging station is constrained by an upper bound due to resource limits. The roads and charging stations are from the real-world data~\citep{blahoudek2020qualitative}.}
    \label{fig:aev}
\end{figure}

In this work, we study CRL from a novel perspective by imposing constraints on \textit{state density functions}, which avoids the use of cost or value function-based constraints in previous CRL literature. Density is a measurement of state concentration in the state space, and is directly related to the state distribution. A variety of real-world constraints are naturally expressed as density constraints in the state space. For example, safety constraints is a special case of density constraints where we require the entire density distribution of states being contained in safe regions. Resource limits can also be encoded by imposing maximum allowed density on certain states. For instance, consider the electric vehicle routing problem~\cite{blahoudek2020qualitative} in Figure~\ref{fig:aev}. In order to keep the remaining energy above a predefined threshold, vehicles can be charged at the charging stations. Due to resource limits, the \emph{vehicle density} at each charging station is constrained by an upper bound. In our experiments, we will demonstrate that density constraints can be used in more general examples, which also include the standard value function-based constraints as special cases.

We cannot directly apply existing CRL optimization methods based on constraining cost or value functions to solve density constrained reinforcement learning (DCRL) problems, because the resulting solution cannot guarantee the satisfaction of required density constraints. Instead,  we propose a novel general model-free algorithm (Algorithm~\ref{alg:primal_dual_template}) that exploits the duality between density functions and Q functions. Such algorithm enables us to incorporate state density constraints straightforwardly in CRL. Our proposed algorithm is applicable of handling both discrete and continuous state and action spaces, and can be flexibly combined with off-the-shelf RL methods to update the policy. We also prove that our algorithm can converge to a near-optimal solution with a bounded error even when the RL-based policy update is imperfect (Theorem~\ref{theo:convergence}).


Comprehensive experiments are conducted on various density constrained problems, which demonstrate the advantages of our approach over state-of-the-art CRL methods in terms of satisfying system specifications and improving the cumulative reward. A notable achievement is that although our method can handle much more general constraints than safety constraints, our DCRL algorithm still shows consistent improvement over existing CRL approaches on standard CRL benchmarks from MuJoCo and Safety-Gym~\cite{Ray2019} where constraints are primarily about safety.

\textbf{Our main contributions are:} 1) We are the first to introduce the DCRL problem with density constraints as a promising way to encode system specifications, which differs from the value function-based constraints in existing CRL literature. 2) We propose a general algorithm for solving DCRL problems by leveraging the duality between density functions and Q functions, and prove that our algorithm is guaranteed to converge to a near-optimal solution with a bounded error even when the policy update is imperfect. 3) We use an extensive set of experiments to examine the advantages of our method over leading CRL approaches, in a wide variety of density constrained tasks as well as standard CRL benchmarks.

\section{Related Work}

Constrained reinforcement learning~\citep{garcia2015comprehensive} primarily focuses on two approaches: modifying the optimality criteria by combining a risk factor~\citep{heger1994consideration, nilim2005robust, howard1972risk, borkar2002q, basu2008learning, sato2001td, dotan2012policy, kadota2006discounted, lotjens2019safe} and incorporating extra knowledge to the exploration process~\citep{moldovan2012safe, abbeel2010autonomous, tang2010parameterized, geramifard2013intelligent, clouse1992teaching, thomaz2006reinforcement, chow2018lyapunov}. 
Our method falls into the first category by imposing constraints and is closely related to constrained Markov decision processes~\citep{altman1999constrained}~(CMDPs). CMDPs has been extensively studied in robotics~\citep{gu2017deep, pham2018optlayer}, game theory~\citep{altman2000constrained}, and communication and networks~\citep{hou2017optimization, bovopoulos1992effect}. Most previous works consider the constraints on value functions, cost functions and reward functions~\citep{altman1999constrained, paternain2019constrained, altman2000constrained, dalal2018safe, achiam2017cpo, ding2020natural}. Instead, we directly impose constraints on the state density functions. Chen et al. \yrcite{chen2019density} and Chen et al. \yrcite{chen2019duality} study the duality between density functions and value functions in CMDPs when the full model dynamics are known, while we consider the model-free reinforcement learning and take a further step to prove the duality of density functions to Q functions. In Geibel and Wysotzki \yrcite{geibel2005risk} density was studied as the probability of entering error states and thus has fundamentally different physical interpretations from us. In Dai et al. \yrcite{dai2017boosting} the duality was used to boost the actor-critic algorithm. The duality is also used in the policy evaluation community \citep{nachum2019dualdice,nachum2020reinforcement,tang2019doubly}. The offline policy evaluation method proposed by Nachum et al. \yrcite{nachum2019dualdice} can also be used to estimate the state density in our paper, but their focus is policy evaluation rather than constrained RL. Therefore, we claim that this paper is the first work to consider density constraints and use the duality property to solve CRL.
\section{Preliminaries}
\label{sec:prelim}

\paragraph{Markov Decision Processes (MDP).}

An MDP $\M$ is a tuple $\langle S, A, P, R, \gamma \rangle$, where
\begin{inparaenum}[(1)]
\item $S$ is the (possibly infinite) set of states;
\item $A$ is the (possibly infinite) set of actions;
\item $P: S \times A \times S \mapsto [0,1]$ is the transition probability with $P(s, a, s')$ the probability of transitioning from state $s$ to $s'$ when action $a \in A$ is taken;
\item $R: S \times A \times S \mapsto \reals$ is the reward associated with the transition $P$ under the action $a \in A$;
\item $\gamma \in [0,1]$ is a discount factor.
\end{inparaenum}

A policy $\pi$ maps states to a probability distribution over actions where $\pi(a|s)$ denotes the probability of choosing action $a$ at state $s$. Let a function $\phi: S \mapsto \reals$ specifie the initial state distribution. 
The objective of an MDP optimization is to find the optimal policy that maximizes the overall discounted reward $J_p = \int_{S}\phi(s)V^{\pi}(s)ds$, where $V^{\pi}(s)$ is called the value function and satisfies $V^{\pi}(s) = r^{\pi}(s) + \gamma \int_{A} \pi(a|s) \int_{S} P(s, a, s') V^{\pi}(s') ds' da$, and $r^{\pi}(s) = \int_{A} \pi(a|s) \int_{S} P(s, a, s') R(s, a, s') ds' da$ is the one-step reward from state $s$ following policy $\pi$. For every state $s$ with occurring as an initial state with probability $\phi(s)$, it incurs a expected cumulative discounted reward of $V^{\pi}(s)$. Therefore the overall reward is $\int_{S}\phi(s)V^{\pi}(s)ds$. The formulation of value functions in MDP typically cannot handle constraints on state distribution, which motivates the density functions.

\paragraph{Density Functions.}
State density functions $\rho: S \mapsto \mathbb{R}_{\geq 0}$ measure the state concentration in the state space~\citep{chen2019duality, rantzer2001a}.\footnote{$\rho$ is not necessarily a probability density function, which means $\int_S \rho(s) = 1$ is not enforced.}
We will show later that generic duality relationship exists between density functions and Q functions, which allows us to directly impose density constraints in RL problems. For infinite horizon MDPs, given a policy $\pi$ and an initial state distribution $\phi$, the stationary density of state $s$ is expressed as:
\begin{align}
    \rho^{\pi}(s) = \sum_{t=0}^{\infty} \gamma^{t} P(s^{t} = s | \pi, s^{0} \sim \phi), \nonumber
\end{align}
which is the discounted sum of the probability of visiting $s$. We prove in the supplementary material that the density has an equivalent expression:
 \begin{align}
     \rho^{\pi}(s) = \phi(s) + \gamma \int_{S} \int_{A} \pi(a|s') P(s', a, s) \rho^{\pi}(s') dads', \nonumber
 \end{align}

\section{Density Constrained Reinforcement Learning}
In this section, we first formally define the novel DCRL problem and clarify its benefits compared to value function-based constraints. To solve DCRL, we will establish the duality between Q functions and density functions, then propose a general algorithm for DCRL with convergence guarantees.

\subsection{Problem Statement}
Given an MDP $\M = \langle S, A, P, R, \gamma \rangle$ and an initial state distribution $\phi$, DCRL finds the optimal policy $\pi^\star$ to the following optimization problem:
\begin{equation}\label{eqn:density_constraint}
    \begin{aligned}
    &\max \int_{S}\phi(s)V^{\pi}(s)ds \\
    &~\mathrm{s.t.}~\rho_{min}(s) \leq \rho^{\pi}(s) \leq \rho_{max}(s), ~\forall s \in S,
    \end{aligned}
\end{equation}
where $\int_{S}\phi(s)V^{\pi}(s)ds$ is the expected cumulative reward. The density constraints $\rho_{min}(s)$ and $\rho_{max}(s)$ are functions of states, and different states can have different lower and upper bound constraints on their densities. 

The formulation in \eqref{eqn:density_constraint} is different from most of the previous work \cite{achiam2017cpo, tessler2019reward, Yang2020Projection}, which did not consider the density constraint but instead used the cost (or reward) value as constraint:
\begin{equation}\label{eqn:cost_constraint}
    \begin{array}{l}
    \max \int_{S}\phi(s)V^{\pi}(s)ds \\
    ~\mathrm{s.t.}~\int_{S}\phi(s)V^{\pi}_C(s)ds \leq \eta,
    \end{array}
\end{equation}
where $V^{\pi}_C$ is the cost value function and $\eta$ is the threshold for the expected cumulative cost.

\paragraph{Benefits of the Density Constraint.} The density constraint in \eqref{eqn:density_constraint} can bring at least two benefits.  First, density has a clear physical and mathematical interpretation as a measurement of state concentration in the state space. A wide range of real-world constraints can be conveniently expressed as density constraints (e.g., the vehicle density in certain areas, and the frequency of agents entering undesirable states). Second, the value function constraint in \eqref{eqn:cost_constraint} requires the time-consuming process of designing and tuning cost functions, which are completely avoided by the density constraint since no cost function tuning is needed. 

\subsection{Duality of Density Functions and Q Functions}
\label{sec:duality}
To solve the DCRL problem in \eqref{eqn:density_constraint}, we cannot directly apply existing RL algorithms \cite{lillicrap2016ddpg, schulman2015trust, schulman2017proximal}, which are designed to optimize the expected cumulative reward or cost, but not the state density function. However, we will show that density functions are actually dual to Q functions, and the density constraints can be realized by modifying the formulation of Q functions. Then the off-the-shelf RL algorithms can be used to optimize the modified Q functions to enforce the density constraints.

We extend the stationary density $\rho^{\pi}$ to consider the action taken at each state. Let $\bar{\rho}:S\times A\to\mathbb{R}_{\geq 0}$ be a stationary state-action density function, which represents the amount of state instances taking action $a$ at state $s$. $\bar{\rho}$ is related to $\rho$ via marginalization: $\rho(s)=\int_A \bar{\rho}(s,a)da$. Under a policy $\pi$, we also have $\bar{\rho}^{\pi}(s,a)=\rho^{\pi}(s)\pi(a|s)$. Let $r(s, a) = \int_S P(s, a, s') R(s, a, s') ds'$. Consider the density function optimization:
\begin{equation}
    \begin{split} \label{eqn:density_constrained_dual}
    &J_{d}=\mathop{\max}\limits_{\bar{\rho},\pi} \int_S\int_A\bar{\rho}^{\pi}(s,a) r(s, a) dads \\
    &\mathrm{s.t.}~\bar{\rho}^{\pi}(s,a)=\pi(a|s) \Big(\phi(s) + \\
    & \qquad \qquad \qquad \gamma \int_S\int_A P(s', a', s)\bar{\rho}^{\pi}(s',a')da'ds'\Big) \\
    &\quad~~ \rho_{min}(s) \leq \rho^{\pi}(s) \leq \rho_{max}(s)
    \end{split}
\end{equation}

\begin{lemma} \label{lemma:same_optimal_policy}
The optimization problems in \eqref{eqn:density_constraint} and \eqref{eqn:density_constrained_dual} share the same optimal policy $\pi^\star$.
\end{lemma}
The proof of Lemma~\ref{lemma:same_optimal_policy} is provided in the supplementary material. It shows that solving \eqref{eqn:density_constraint} is equivalent to solving \eqref{eqn:density_constrained_dual}. Now we are ready to present the duality between density functions and Q functions, which reveals that \eqref{eqn:density_constrained_dual} can be solved via standard RL algorithms by modifying the Q function. We use the Lagrangian method for \eqref{eqn:density_constrained_dual} and denote the Lagrange multipliers for $\rho^{\pi} \ge \rho_{min}$ and $\rho^{\pi}\le\rho_{max} $ as $\sigma_{-}:S \mapsto \mathbb{R}_{\geq 0}$ and $\sigma_{+}: S \mapsto \mathbb{R}_{\geq 0}$. Then we consider the following optimization problem:
\begin{equation} \label{eqn:adjusted_q_optimization}
    \begin{split}
    &J_{p} = \mathop{\max}\limits_{Q,\pi} \int_S \phi(s) \int_A Q^{\pi}(s,a)\pi(a|s) da ds \\
    &\mathrm{s.t.}~ Q^{\pi}(s,a)= r(s, a) + \sigma_{-}(s) - \sigma_{+}(s) + \\
    & \qquad \quad \gamma \int_S P(s, a, s') \int_A \pi(a'|s') Q^{\pi}(s',a') da'ds'
    \end{split}
\end{equation}
The difference between the Q function in \eqref{eqn:adjusted_q_optimization} and the standard Q function is that the reward $r(s, a)$ is modified to $r(s, a) + \sigma_{-}(s) - \sigma_{+}(s)$.
\begin{theorem} \label{theo:density_constrained}
	The density constrained optimization objectives $J_{d}$ in \eqref{eqn:density_constrained_dual} and $J_{p}$ in \eqref{eqn:adjusted_q_optimization} are dual to each other. If both are feasible and the KKT conditions are satisfied, then they share the same optimal policy $\pi^\star$.
\end{theorem}

The proof of Theorem~\ref{theo:density_constrained} is provided in the appendix. Theorem~\ref{theo:density_constrained} reveals that when the KKT conditions are satisfied, the optimal solution to the adjusted unconstrained primal problem (\ref{eqn:adjusted_q_optimization}) is exactly the optimal solution to the dual problem (\ref{eqn:density_constrained_dual}) with density constraints. Such an optimal solution not only satisfies the state density constraints, but also maximizes the the total reward $J_d$. Thus it is the optimal solution to the DCRL problem. Note that \eqref{eqn:adjusted_q_optimization} is solvable using standard RL algorithms \cite{lillicrap2016ddpg, schulman2015trust, schulman2017proximal}. However, the update of policy $\pi$ will also result in an update of state density function $\rho^{\pi}$, which will change the Lagrangian multipliers $\sigma_+$ and $\sigma_-$ to in order to satisfy the KKT conditions of \eqref{eqn:density_constrained_dual}. This motivates us to develop an algorithm that iteratively update $\pi, \rho^{\pi}, \sigma_+ \text{~and~} \sigma_-$, which will be detailed in Section~\ref{sect:algorithm}.

\textbf{Remark 1.} While the duality between density functions and \emph{value functions} has already been studied \cite{chen2019duality}, we take a step further to show the duality property between the density functions and \emph{Q functions}. We prove and use the duality between density functions and Q functions over continuous state space to solve density constrained RL from a novel perspective, which has not been explored and utilized by published RL literature. 

\textbf{Remark 2.} The dual variables in Linear Programming are different from ours and do not have a clear physical interpretation. Technically, any non-negative dual variable satisfying the conservation law (Liouville PDE in the continuous case, see Chen et al. \yrcite{chen2019duality}) is a valid dual. However, among all valid dual variables, the state density is associated with a clear physical interpretation as the concentration of states, and we are able to directly apply constraints on the density in RL. 

\textbf{Remark 3.} The value function-based constraint in CMDPs can be viewed as a special case of density constraint. $\int_{S}\phi(s)V^{\pi}_C(s)ds \leq \eta$ is equivalent to $\int_S \rho^{\pi}(s)r_c(s)ds \leq \eta$. Thus the value function-based CRL problems can also be solved by the DCRL framework.

\textbf{Remark 4.} The general DCRL problem cannot be solved by value function-based RL methods. One may argue that for a single state $s_k$, its visiting frequency can be constrained by defining a cost function $c_k$ that returns 1 when $s_k$ is visited and 0 otherwise. Then value function-based methods can be used to constrain the expected value $V_{C_k}$ of the cost, as is in $\int_{S}\phi(s)V_{C_k}(s)ds \leq \eta_k~(*)$. However, when every state $s_j$ has unique density constraints $\rho_{min}(s_j)$ and $\rho_{max}(s_j)$, every state $s_j$ needs a value function $V_{C_j}$ (for the cost of visiting $s_j$) and an inequality  $(*)$. In order to cover all states, it would require an extremely large and even infinite number (when the state space is continuous) of value functions and inequalities, which can make the problem intractable for value function-based methods. Furthermore, we have explicitly mentioned in Remark 3 that the value function-based constraint in CMDP is equivalent to a special case of density constraint, and our general formulation of state-wise density constraint is more expressive and cannot be trivially converted into value function-based constraints.

\subsection{The DCRL Algorithm}
\label{sect:algorithm}

\begin{figure*}
    \centering
    \includegraphics[width=0.9\linewidth]{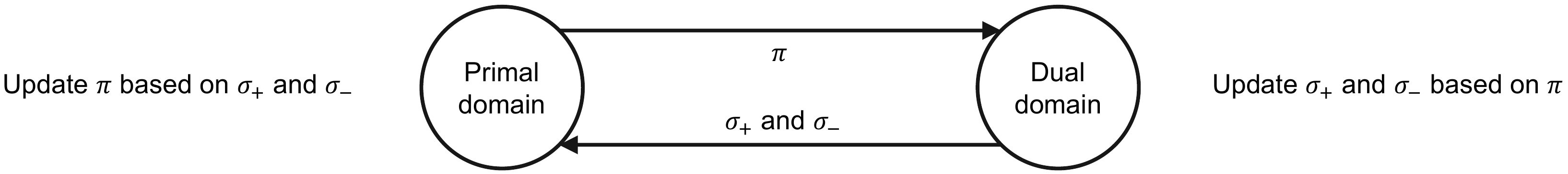}
    \caption{\footnotesize Illustration of the iterative optimization in DCRL. In the primal domain, we solve the adjusted primal problem in (\ref{eqn:adjusted_q_optimization}) to obtain the policy $\pi$. Then in the dual domain, the $\pi$ is used to evaluate the state density. The Lagrange multipliers $\sigma_{+}$ and $\sigma_{-}$ are updated as $\sigma_{+} \leftarrow \max(\mathbf{0}, \sigma_{+} + \alpha(\rho^{\pi} - \rho_{max}))$ and $\sigma_{-} \leftarrow \max(\mathbf{0}, \sigma_{-} + \alpha(\rho_{min} - \rho^{\pi}))$. In the next loop, since the reward $r(s, a) + \sigma_{-}(s) - \sigma_{+}(s)$ is updated, the primal optimization solves for the new $\pi$ under the updated reward. The loop stops when the KKT conditions are satisfied.}
    \label{fig:primal_dual}
\end{figure*}
The density function optimization in continuous state space is an infinite dimensional linear program and there is no convenient approximation method. We thus turn to the equivalent Q function optimization where standard RL tools are applicable.
By utilizing the duality between density function and Q function (Theorem~\ref{theo:density_constrained}) in the density constrained optimization, the DCRL problem can be solved by alternating between the primal problem \eqref{eqn:adjusted_q_optimization} and dual problem \eqref{eqn:density_constrained_dual}, as is illustrated in Figure~\ref{fig:primal_dual}. In the primal domain, we solve the adjusted primal problem (reward adjusted by Lagrange multipliers) in (\ref{eqn:adjusted_q_optimization}) using off-the-shelf unconstrained RL methods such as TRPO~\citep{schulman2015trust} and DDPG~\citep{lillicrap2016ddpg}. Note that the density constraints are enforced in dual domain and the primal domain is still an unconstrained problem, which means we can make use of existing RL methods to solve the primal problem. In the dual domain, the policy is used to evaluate the state density function, which is described in details in Section~\ref{sec:computational_approaches}. If the KKT conditions $\sigma_{+} \cdot(\rho^\pi - \rho_{max})=0, ~ \sigma_{-}\cdot(\rho_{min} - \rho^\pi) = 0$ and $\rho_{min} \leq \rho^{\pi} \leq \rho_{max}$ are not satisfied, the Lagrange multipliers are updated and enter the next loop. The key insight is that the density constraints can be enforced in the dual problem, and we can solve the dual problem by solving the equivalent primal problem using existing algorithms. Alternating between the primal and dual optimization can gradually adjust the Lagrange multipliers until the KKT conditions are satisfied.

\label{sec:algorithm}
\begin{algorithm}
    \caption{Template of the DCRL algorithm}
    \begin{algorithmic}[1]
        \STATE \textbf{Input} MDP $\M$, initial condition distribution $\phi$, constraints on the state density $\rho_{max} $ and $\rho_{min}$
        \STATE Initialize $\pi$ randomly, $\sigma_{+} \leftarrow \mathbf{0}$, $\sigma_{-} \leftarrow \mathbf{0}$
        \STATE Generate experience $D_{\pi} \subset \{(s, a, r, s') ~|~ s^{0} \sim \phi, a\sim\pi(s) \text{~then~} r \text{~and~} s' \text{~are observed}\}$
        \REPEAT
        \STATE $(s, a, r, s') \leftarrow (s, a, r + \sigma_{-}(s) - \sigma_{+}(s), s')$ for each $(s, a, r, s')$ in $D_{\pi}$ \label{alg:update_Dpi}
        \STATE Solve for $\pi$ of \eqref{eqn:adjusted_q_optimization} using $D_{\pi}$ via standard RL \label{line:alg1_update_policy}
        \STATE Generate experience $D_{\pi} \subset \{(s, a, r, s') ~|~ s^{0} \sim \phi, a\sim\pi(s) \text{~then~} r \text{~and~} s' \text{~are observed}\}$ \label{alg:experiences}
        \STATE Compute stationary density $\rho^{\pi}$ using $D_{\pi}$ \label{line:alg1_comp_density}
        \STATE $\sigma_{+} \leftarrow \max(\mathbf{0}, \sigma_{+} + \alpha(\rho^{\pi} - \rho_{max}))$ \label{line:alg1_update_sigma_max}
        \STATE $\sigma_{-} \leftarrow \max(\mathbf{0}, \sigma_{-} + \alpha(\rho_{min} - \rho^{\pi}))$ \label{line:alg1_update_sigma_min}
        \UNTIL{$\sigma_{+}\cdot(\rho^\pi-\rho_{max})=0,~ \sigma_{-}\cdot(\rho_{min}-\rho^\pi)=0~ \AND~ \rho_{min} \leq \rho^{\pi} \leq \rho_{max}$}
        \STATE \textbf{Return} $\pi,~\rho^{\pi}$
    \end{algorithmic}
    \label{alg:primal_dual_template}
\end{algorithm}

A general template of the density constraint policy optimization is provided in Algorithm~\ref{alg:primal_dual_template}. In Algorithm~\ref{alg:primal_dual_template}, the Lagrange multipliers $\sigma_{+}$ and $\sigma_{-}$ are used to adjust rewards, which lead to an update of the policy $\pi$. Then the policy is used to evaluate the stationary density, and the Lagrange multipliers are updated following dual ascent. The iteration stops when all the KKT conditions are satisfied.

\subsection{Convergence Guarantee under Imperfect Updates}
If every loop of Algorithm~\ref{alg:primal_dual_template} gives the perfect (optimal) policy $\pi$ of \eqref{eqn:adjusted_q_optimization} under the current $\sigma_-$ and $\sigma_+$, then the convergence can be derived from the subgradient methods \cite{boyd2003subgradient} and will not be detailed here. But in reality, the policy update in each loop can be imperfect, which is unavoidable for standard RL algorithms. 

In light of this, we are interested in the case where the policy solved at each iteration is imperfect, and will provide convergence guarantees of \alg{\ref{alg:primal_dual_template}}. We will also identify how the error in policy updates propagates to the solution found by Algorithm~\ref{alg:primal_dual_template} and derive an upper bound of its distance to the optimal solution. Denote the negative objective function of \eqref{eqn:density_constrained_dual} as $f(\rho^{\pi}) = -\int_S\int_A\rho^{\pi}(s)\pi(a|s) r(s, a) dads$. We assume $f$ is strongly convex modulus $\mu$. If this is not the case, one can add a strongly convex regularization term to the (already convex) objective function. Then \eqref{eqn:density_constrained_dual} minimizes $f(\rho^{\pi})$ subject to the density constraints. For brevity, we only consider the density upper bound in this convergence analysis, and the same result exists for the density lower bound. The Lagrangian dual function is formulated as
\begin{equation} \label{eqn:d_sigma}
    \begin{aligned}
    d(\sigma_{+}) = \mathop{\min}_{\rho^{\pi}} f(\rho^{\pi}) + \int_S \sigma_{+}(s) (\rho^{\pi}(s) - \rho_{max}(s))ds 
    \end{aligned}
\end{equation}
Denote the feasible set of Lagrangian multipliers as $\mathcal{P}$. The dual ascent is to find $\mathop{\max}_{\sigma_{+}\in \mathcal{P}} d(\sigma_{+})$. We assume the policy update is imperfect and the suboptimality of the solution $\hat{\rho}^{\pi}$ to \eqref{eqn:density_constrained_dual} is upper bounded as:
\begin{equation}
    f(\hat{\rho}^{\pi}) + \int_S \sigma_{+}(s) (\hat{\rho}^{\pi}(s) - \rho_{max}(s))ds - d(\sigma_{+}) \leq \epsilon \nonumber
\end{equation}
And the imperfect dual ascent takes the form $\sigma_{+} \leftarrow \max(\mathbf{0}, \sigma_{+} + \alpha \nabla \hat{d}(\sigma_+))$, where $\nabla\hat{d}(\sigma_+) = \hat{\rho}^{\pi} - \rho_{max}$. Let $\hat{g}(\sigma_{+}, \alpha) = \frac{1}{\alpha} (\max(\mathbf{0}, \sigma_{+} + \alpha \nabla \hat{d}(\sigma_{+})) - \sigma_{+})$, which is the residual of $\sigma_+$ before and after the imperfect update in a single loop of Algorithm~\ref{alg:primal_dual_template}.
\begin{lemma} \label{lemma:imperfect_dual_ascent}
Following the imperfect dual ascent with step size $\alpha \leq \mu$, we have
\begin{equation}
    d(\sigma_{+}^{k+1}) \geq d(\sigma_{+}^{k}) + \frac{\alpha}{2} ||\hat{g}(\sigma_{+}^k, \alpha)||^2 - \sqrt{\frac{2\epsilon}{\mu}} ||\hat{g}(\sigma_{+}^k, \alpha)|| \nonumber
\end{equation}
\end{lemma}
The superscript $k$ denotes the $k^{th}$ loop of Algorithm~\ref{alg:primal_dual_template}. The proof of Lemma~\ref{lemma:imperfect_dual_ascent} is provided in the supplementary material. Let $\mathcal{P}^{\star} = \{\sigma_+| d(\sigma_+) = \mathop{\max}_{\sigma_{+}\in \mathcal{P}} d(\sigma_{+})\}$ be the set of optimal solutions to \eqref{eqn:d_sigma}.
\begin{theorem} \label{theo:convergence}
    There exists constants $\lambda > 0$ and $\xi > 0$ such that Algorithm~\ref{alg:primal_dual_template} with imperfect policy updates converges to a dual solution $\hat{\sigma}_+$ that satisfies
    \begin{equation}
    \min_{\sigma_+'\in \mathcal{P}^{\star}} ||\hat{\sigma}_+(s) - \sigma_+'|| \leq \lambda \sqrt{\frac{\epsilon}{\mu}}
    \end{equation}
    The dual function also converges to a bounded neighbourhood of its optimal value:
    \begin{equation}
        \min_{\sigma_+'\in \mathcal{P}^{\star}} ||d(\hat{\sigma}_+) - d(\sigma_+')|| \leq  \xi \frac{\epsilon}{\mu^2}
    \end{equation}
\end{theorem}

The proof of Theorem~\ref{theo:convergence} is provided in the supplementary material. Theorem~\ref{theo:convergence} shows that Algorithm~\ref{alg:primal_dual_template} is guaranteed to converge to a near-optimal solution even under imperfect policy updates.

\subsection{Computational Approaches}
\label{sec:computational_approaches}
\alg{\ref{alg:primal_dual_template}} requires computing the policy $\pi$, density $\rho^{\pi}$, Lagrange multipliers $\sigma_{+}$ and $\sigma_{-}$. For $\pi$, there are well-developed representations such as neural networks and tabular methods. Solving $\pi$ from experience $D_{\pi}$ is also straightforward via standard model-free RL. By contrast, the computation of $\rho^{\pi}$, $\sigma_{+}$ and $\sigma_{-}$ need to be addressed. 

\textbf{Density Functions.} In the discrete state case, $\rho^{\pi}$ is represented by a vector where each element corresponds to a state. To compute $\rho^{\pi}$ from experience $D_{\pi}$ (line~\ref{line:alg1_comp_density} of \alg{\ref{alg:primal_dual_template}}), let $D_{\pi}$ contain $N$ episodes where episode $i$ ends at time $T_i$. Let $s^{ij}$ represent the state reached at time $j$ in the $i^{th}$ episode. Initialize $\rho^{\pi} \leftarrow \mathbf{0}$. For all $i \in \{1, \cdots, N\}$ and $j \in \{0, 1, \cdots, T_i\}$, do the update $\rho^{\pi}(s^{ij}) \leftarrow \rho^{\pi}(s^{ij}) + \frac{1}{N}\gamma^{j}$.
The resulting vector $\rho^{\pi}$ approximates the stationary state density. In the continuous state space, $\rho^{\pi}$ cannot be represented as a vector since there are infinitely many states. We utilize the kernel density estimation method~\citep{chen2017tutorial,chen2019duality} that computes $\rho^{\pi}(s)$ at state $s$ using the samples in $D_{\pi}$ with $\rho^{\pi}(s) = \frac{1}{N} \sum_{i=1}^{N}\sum_{j=0}^{T_i} \gamma^{j} K_{h}(s - s^{ij})$ 
where $K_{h}$ is the kernel function satisfying $\forall s\in S, K_{h}(s) \geq 0$ and $\int_{S}K_{h}(s)ds=1$. There are multiple choices of the kernel $K_{h}$, e.g. Gaussian, Spheric, and Epanechnikov kernels~\cite{chen2017tutorial}, and probabilistic guarantee of accuracy can be derived \citep{kerneldensitycmu}.

\textbf{Lagrange Multipliers.} If the state space is discrete, both $\sigma_{+}$ and $\sigma_{-}$ are vectors whose length equals to the number of states. In each loop of \alg{\ref{alg:primal_dual_template}}, after the stationary density is computed, $\sigma_{+}$ and $\sigma_{-}$ are updated following Line~\ref{line:alg1_update_sigma_max} and Line~\ref{line:alg1_update_sigma_min} respectively in Algorithm~\ref{alg:primal_dual_template}. If the state space is continuous, we construct Lagrange multiplier functions $\sigma_{+}$ and $\sigma_{-}$ from samples in the state space leveraging linear interpolation. Let $\bar{s} = \left[s_1, s_2, \cdots \right]$ represent the samples in the state space. In every loop of \alg{\ref{alg:primal_dual_template}}, denote the Lagrange function computed by the previous loop as $\sigma_{+}^{o}$ and $\sigma_{-}^{o}$. We compute the updated Lagrange multipliers at states $\bar{s}$ as $\bar{\sigma}_{+} = [\max (0, \sigma_{+}^{o}(s_1) + \alpha( \rho^{\pi}(s_1) - \rho_{max}(s_1)), \max (0, \sigma_{+}^{o}(s_2) + \alpha(\rho^{\pi}(s_2) - \rho_{max}(s_2)), \cdots ]$ and $\bar{\sigma}_{-} = [\max (0, \sigma_{-}^{o}(s_1) + \alpha(\rho_{min}(s_1) - \rho^{\pi}(s_1)), \max (0, \sigma_{-}^{o}(s_2) + \alpha(\rho_{min}(s_2) - \rho^{\pi}(s_2)), \cdots ]$. Then the new $\sigma_{+}$ and $\sigma_{-}$ are obtained by linearly interpolating $\bar{\sigma}_{+}$ and $\bar{\sigma}_{-}$ respectively.
\section{Experiment}
\label{sec:experiments}

We consider a wide variety of CRL benchmarks and demonstrate how they can be effectively solve by the proposed DCRL approach. The density constrained benchmarks include autonomous electrical vehicle routing~\cite{blahoudek2020qualitative}, safe motor control~\cite{traue2019reinforcement} and agricultural drone control. The standard CRL benchmarks are from MuJoCo and Safety-Gym~\cite{Ray2019}. The definition of reward and constraint vary from task to task and will be explained when each task is introduced.

\paragraph{Baseline Approaches.}
Three CRL baselines are compared. \textbf{PCPO}~\citep{Yang2020Projection} first performs an unconstrained policy update then project the action to the constrained set. \textbf{CPO}~\citep{achiam2017cpo} maximizes the reward in a small neighbourhood that enforces the constraints. \textbf{RCPO}~\citep{tessler2019reward} incorporates the cost terms and Lagrange multipliers with the reward function to encourage the satisfaction of the constraints. We used the original implementation of CPO and PCPO with KL-projection that leads to the best performance. For RCPO, since the official implementation is not available, we re-implemented RCPO and made sure it matches the original performance. All the three baseline approaches and our DCRL have the same number of neural network parameters. Note that the baseline approaches enforce the constraints by restricting \emph{values functions} while our DCRL restricts the \emph{state density functions}. When we train the value function-based CRL methods on density constrained tasks, we convert the density threshold to the value function threshold by duality between density functions and value functions~\citep{chen2019duality}.

\subsection{Autonomous Electrical Vehicle Routing}

\begin{figure}[ht]
    \centering
    \includegraphics[width=\linewidth]{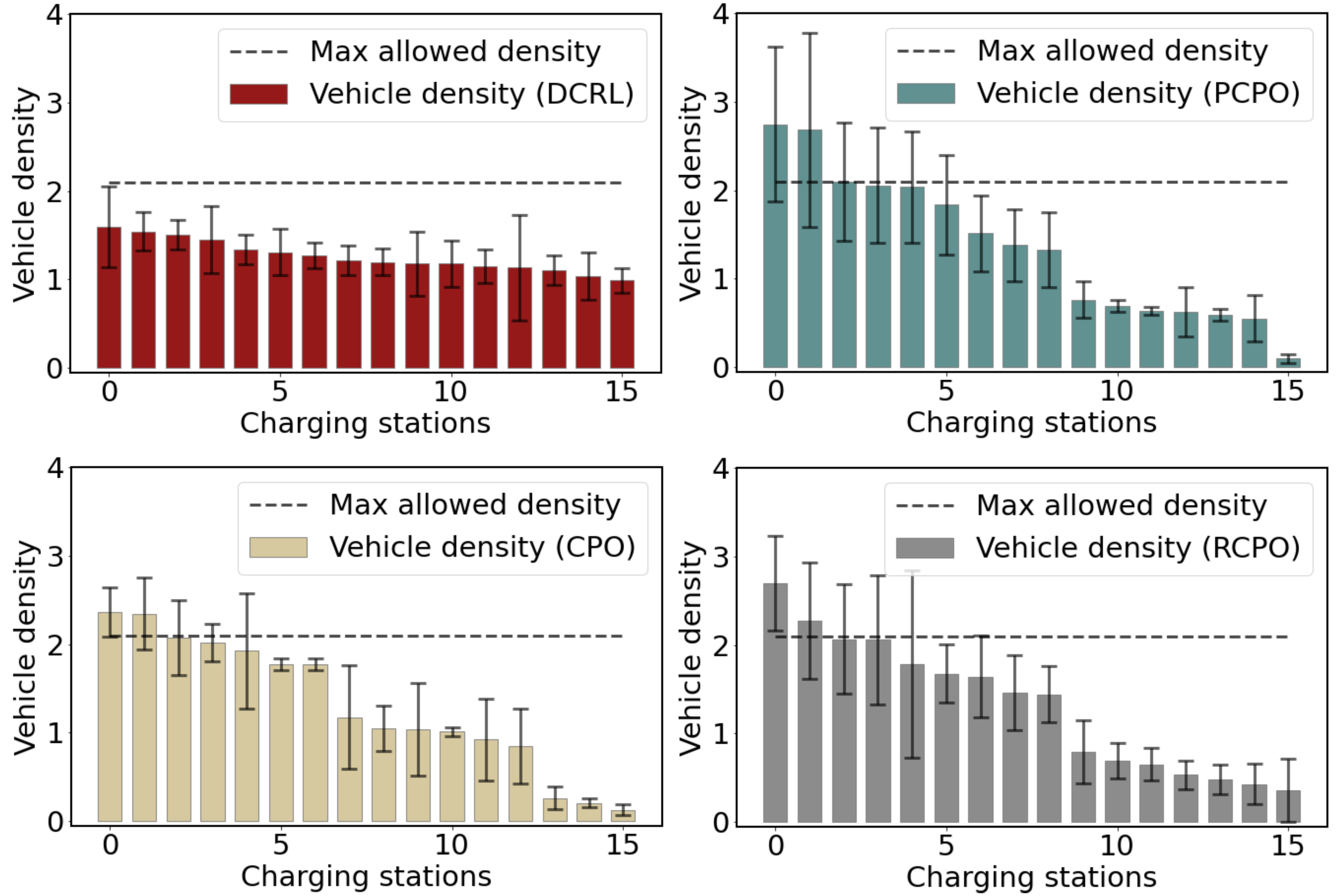}
    \caption{\footnotesize Vehicle densities at the charging stations. All results are averaged over 10 independent trials. 16 charging stations with the highest vehicle densities are kept for visualization.}
    \label{fig:aev_density}
\end{figure}

\begin{figure}[ht]
    \centering
    \includegraphics[width=\linewidth]{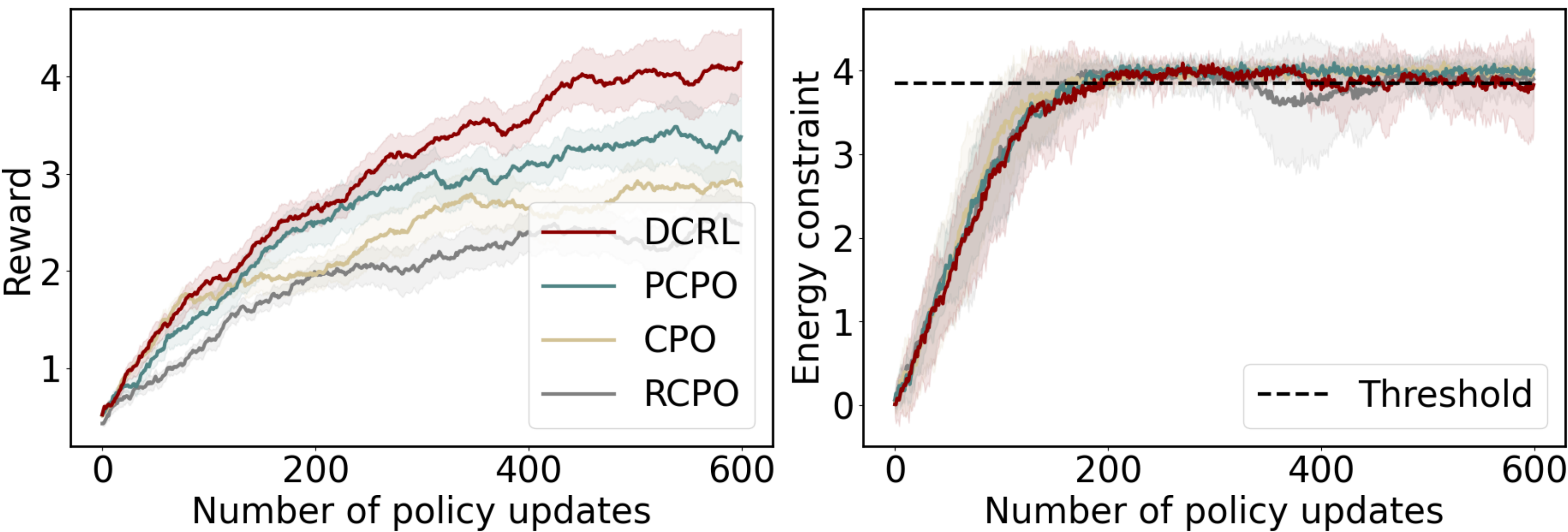}
    \caption{\footnotesize Average reward and remaining energy. All results are averaged over 10 independent trials.}
    \label{fig:aev_reward}
\end{figure}

\begin{figure*}[t]
    \centering
    \includegraphics[width=\linewidth]{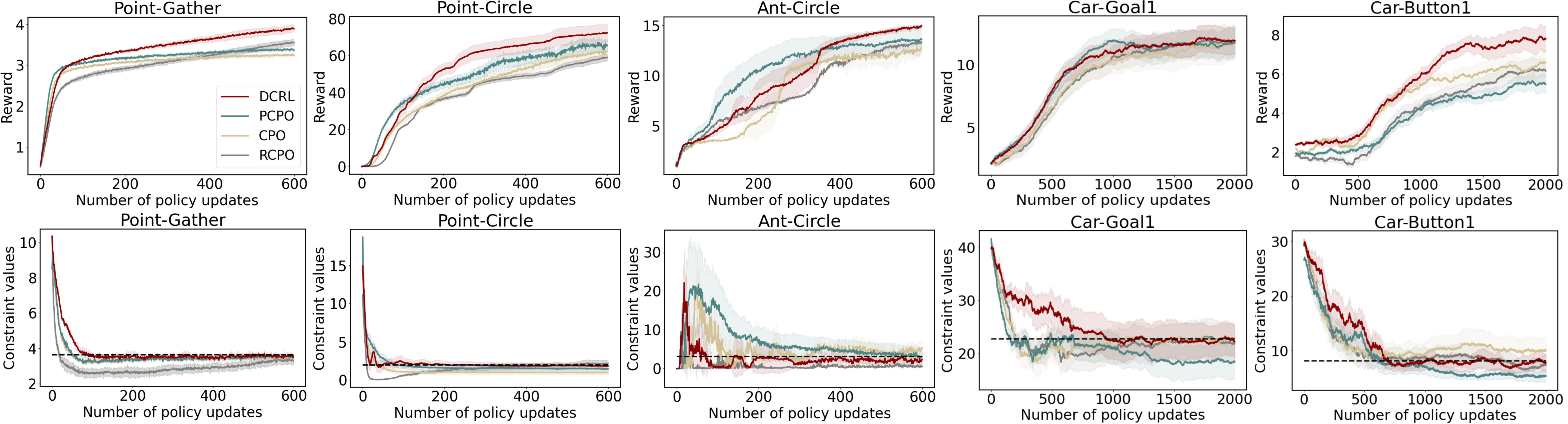}
    \caption{\footnotesize Performance on the constrained reinforcement learning tasks on the MuJoCo~\citep{todorov2012mujoco} simulator. The first three tasks are from CPO~\cite{achiam2017cpo} and the last two tasks are from the Safety-Gym~\cite{Ray2019}. All results are averaged over 10 independent trials. The constraint thresholds are all upper bounds.}
    \label{fig:mujoco_result}
\end{figure*}

Our first case study is about controlling autonomous electric vehicles (EV) in the middle of Manhattan, New York. It is adopted from  Blahoudek et al. \yrcite{blahoudek2020qualitative} and is shown in Figure~\ref{fig:aev}.
While EVs drive to their destinations, they can avoid running out of power by recharging at the fast charging stations along the roads. At the same time, the vehicles should not stay at the charging stations for too long in order to save resources and avoid congestion. An road intersection is called a node. In each episode, an autonomous EV starts from a random node and drives to the goals. At each node, the EV chooses a direction and reaches the next node along that direction at the next step. The consumed electric energy is assumed to be proportional to traveling distance. There are 1024 nodes and 137 charging stations in total. Denote the full energy capacity of the vehicle as $c_f$ and the remaining energy as $c_r$. When arriving at a charging station, the EV chooses a charging time $\tau \in \left[0, 1\right]$, then its energy increases to $\min (c_f, c_r + \tau c_f)$. The action space includes the EV direction and the charging time $\tau$. The state space $S \subset \mathbb{R}^3$ is consisted of the current 2D location and the remaining energy $c_r$. The agent receives a negative reward proportional to its traveling distance to restrict energy consumption, and a +10 reward when it reaches the goal. In one episode, a single agent starts from a random initial location and reaches the goal. The density is accumulated for 2000 episodes, which is equivalent to having 2000 agents.

Two types of constraints are considered: (1) the minimum remaining energy should keep close to a required threshold and (2) the vehicle density at charging stations should be less than a given threshold. Apparently, if the EV chooses a larger $\tau$, then the constraint (1) is more likely to be satisfied, while (2) is more likely to be violated, since a larger $\tau$ will increase the vehicle density at the charging station. These contradictory constraints pose a greater challenge to the RL algorithms. Both constraints can be naturally expressed as density constraints. For constraint (1), we can limit the density of low-energy states. For constraint (2), it is straightforward to limit the EV density (a function of $\mathbb{E}[\tau]$) at charging stations. The threshold of density constraints are transformed to the threshold of value functions to be used by the baseline methods. The conversion is based on the duality of density functions and value functions~\citep{chen2019duality}. Figure~\ref{fig:aev_density} demonstrates that our DCRL can avoid the vehicle densities from exceeding the thresholds, while the baseline methods suffer from constraint violation. This is because DCRL allows us to explicitly set state density thresholds. Figure~\ref{fig:aev_reward} shows that all the compared methods can satisfy the constraint on remaining energy, and DCRL can reach the highest reward.

\subsection{MuJoCo Benchmark and Safety-Gym}

Experiments are conducted on three tasks adopted from CPO~\citep{achiam2017cpo} and two tasks from the Safety-Gym~\cite{Ray2019}, built on top of the MuJoCo simulator. The tasks include Point-Gather, Point-Circle, Ant-Circle, Car-Goal and Car-Button. In the Point-Gather task, a point agent moves on a plane and tries to gather as many apples as possible while keeping the probability of gather bombs below the threshold. In the Point-Circle task, a point agent tries to follow a circular path that maximizes the reward, while constraining the probability of exiting an given area. The Ant-Circle task is similar to the Point-Circle task except that the agent is an ant instead of a point. Detailed configurations of the three benchmarks can be found in Achiam et al.~\yrcite{achiam2017cpo}. In the Car-Goal task, a car agent tries to navigate to a goal while avoiding hazards. In the Car-Button task, a car agent tries to press the goal button while avoiding hazards, and not to press the wrong buttons. Detailed descriptions of the Safety-Gym benchmark can be found in Ray et al.~\yrcite{Ray2019}. The original value function constraints for these benchmarks are converted to state density constraints for DCRL by duality between density functions and value functions.

Figure~\ref{fig:mujoco_result} demonstrates the performance of the four methods. In general, DCRL is able to achieve higher reward than other methods while satisfying the constraint thresholds. In the Point-Gather and Point-Circle environments, all the four approaches exhibit stable performance with relatively small variances. In the Ant-Circle environment, the variances of reward and constraint values are significantly greater than that in Point environments, which is mainly due to the complexity of ant dynamics. In Ant-Circle, after 600 iterations of policy updates, the constraint values of the four approaches converge to the neighbourhood of the threshold. The reward of DCRL falls behind PCPO in the first 400 iterations of updates but outperforms PCPO thereafter.

\begin{figure}[t]
    \centering
    \includegraphics[width=\linewidth]{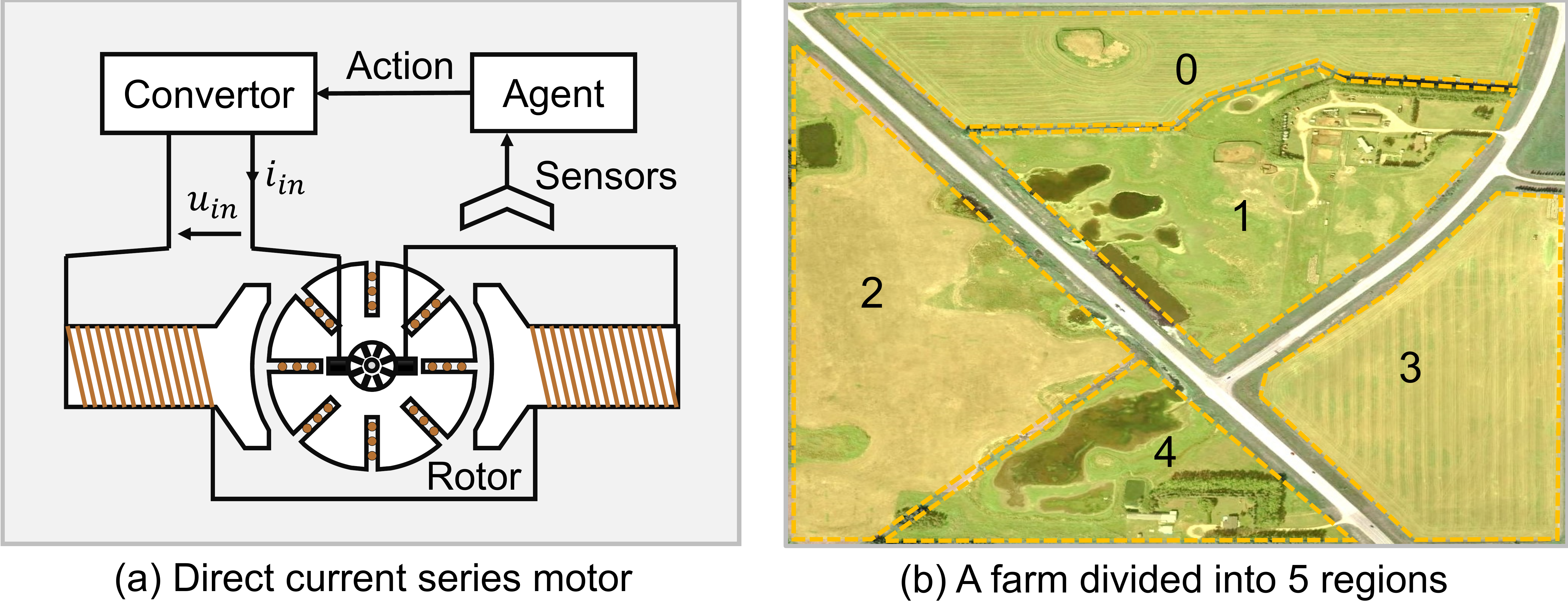}
    \caption{\footnotesize Illustration of the safe motor control and drone application environments. (a) Control the rotor to follow a reference angular velocity trajectory while limiting the density of high-temperature states. (b) Control the drones to spray pesticide over a  farmland which is divided into five parts and each requires different densities of pesticide.}
    \label{fig:motor_and_farm_env}
\end{figure}

\begin{figure}
    \centering
    \includegraphics[width=\linewidth]{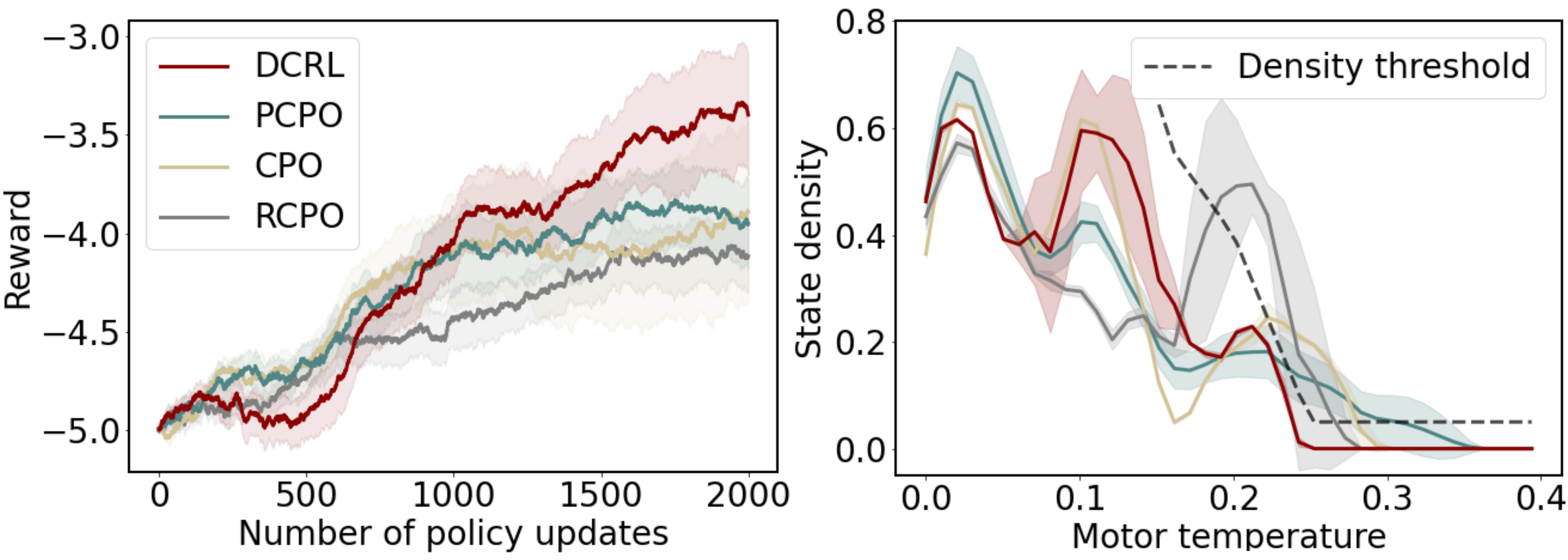}
    \caption{\footnotesize Reward and state density constraint in the safe motor control task. Results are averaged over 10 independent trials.}
    \label{fig:motor_reward_icml}
\end{figure}

\begin{figure}[t]
    \centering
    \includegraphics[width=\linewidth]{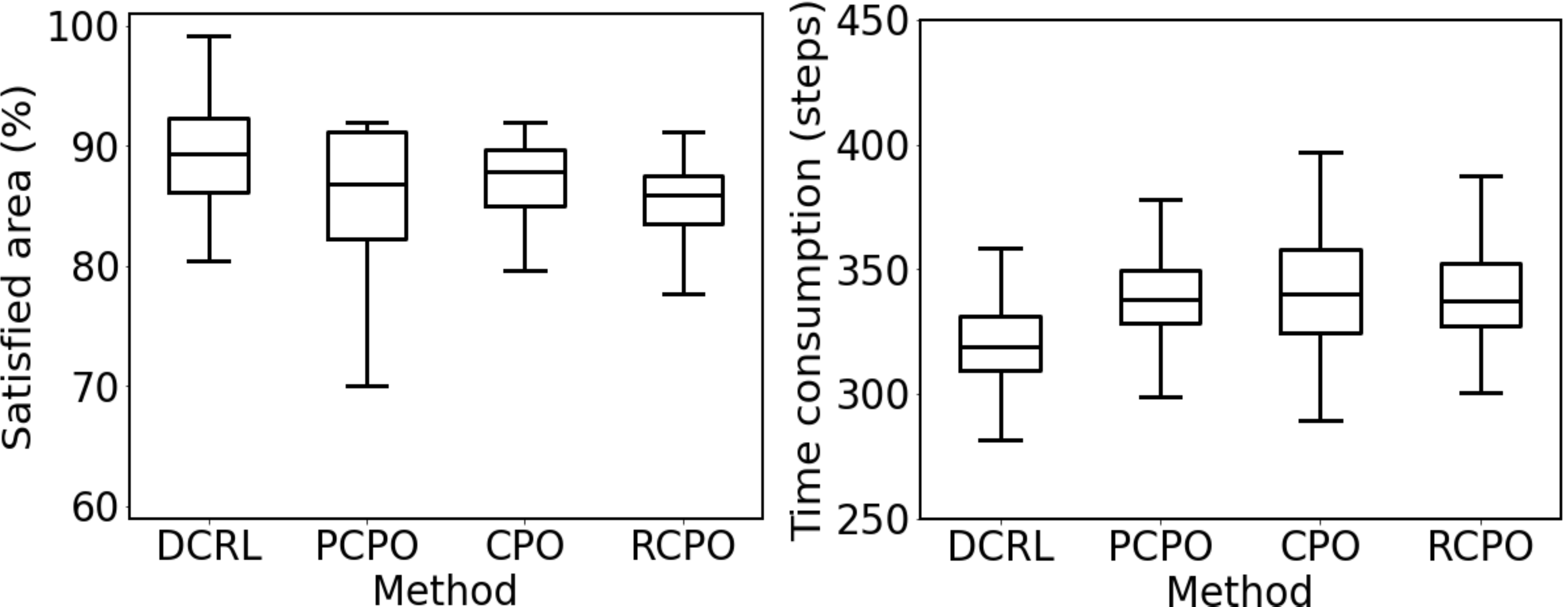}
    \caption{\footnotesize Result of the agricultural spraying problem. (a) Percentage of the entire area that satisfies the pesticide density requirement. (b) Time consumption in steps to reach the destination. Whiskers in the plots denote confidence intervals. Results are calculated with 10 independent trials after each method converges.}
    \label{fig:farm_result}
\end{figure}

\subsection{Safe Motor Control}

The safe motor control environment is adopted from Traue et al.~\yrcite{traue2019reinforcement} and shown in Figure~\ref{fig:motor_and_farm_env}~(a). The objective is to control the direct current series motor and ensure its angular velocity follows a random reference trajectory and prevent the motor from overheating. The state space $S \subset \mathbb{R}^6$ and consists of six variables: angular velocity, torque, current, voltage, temperature and the reference angular velocity. The action space $A \subset \mathbb{R}$ is the electrical duty cycle that controls the motor power. The agent outputs a duty cycle at each time step to drive the angular velocity close to the reference. When the reference angular velocity increases, the required duty cycle will need to increase. As a result, the motor's power and angular velocity will increase and cause the motor temperature to grow. The reward is defined as the negative distance between the measured angular velocity and the reference angular velocity. We consider the state density w.r.t. the motor temperature as constraint. High-temperature states should have a low density to protect the motor. When we train the baseline methods, the density constraints are converted to equivalent cost value constraints by duality between density functions and value functions. Figure~\ref{fig:motor_reward_icml} shows the reward and state density for each method. We find that DCRL is successful at reaching the highest reward among the compared methods while respecting the state density constraints. More experiments under different settings can be found in the supplementary material.

\subsection{Agricultural Spraying Drone}
We consider the problem of using drones to spray pesticides over a farmland in simulation. The farmland is shown in Figure~\ref{fig:motor_and_farm_env}~(b), which is divided into 5 regions and different regions require different pesticide densities represented by a lower bound $\rho_{min}$ and an upper bound $\rho_{max}$. The drone starts from the top-left corner, flies over the farmland spraying pesticides, and stops at the bottom-right corner. The state space constrains the position and velocity of the drone, as well as the required pesticide density of the current region. The action space contains the angular acceleration and the vertical thrust to control the drone. The drone sprays a constant volume of pesticide at each time step, and controls the pesticide density on the land by adjusting its position and velocity. A +1 reward will be given when the pesticide density of the area below the drone enters the required range, and a -1 reward will be given when the density exists the range. A +10 reward will be given when the drone reaches the bottom-right corner that is the destination. Figure~\ref{fig:farm_result} shows the experimental results. With the DCRL method, we can obtain the highest percentage of area that satisfies the pesticide density constraint. Also, DCRL requires the least steps to reach the destination. Details of the experiment configurations and results under different settings can be found in the supplementary material.

\section{Conclusion and future works}

We study a novel class of constrained reinforcement learning problems where the constraints are imposed on state density functions, rather than value functions considered by previous literature. State densities have clear physical meanings and can express a variety of constraints of the environment and the system. We prove the duality between density functions and Q functions, then leverage the duality to develop a general algorithm to solve the DCRL problems. We also provide the convergence guarantee of our algorithm even under imperfect policy updates. 
Note that our algorithm does not guarantee the satisfaction of density constraints during training, and we do see occasional violations at the beginning iterations. In the future, we aim to improve the algorithm to enforce the density constraints throughout the entire training process. 

\section{Acknowledgement}
The authors acknowledge support from the DARPA Assured Autonomy under contract FA8750-19-C-0089 and from the Defense Science and Technology Agency in Singapore. The views, opinions, and/or findings expressed are those of the authors and should not be interpreted as representing the official views or policies of the Department of Defense, the U.S. Government, DSTA Singapore, or the Singapore Government.

\bibliography{reference}
\bibliographystyle{icml2021}

\clearpage
\appendix
\section{Proofs of Statements and Theorems}
\subsection{Equivalent Expression of the Density Function}
In Section~\ref{sec:prelim}, we point out that the state density function $\rho^{\pi}$ has two equivalent expressions, which we prove as follows:
\begin{align}
	\rho^{\pi}(s) 
	& = \sum_{t=0}^{\infty} \gamma^{t} P(s^{t} = s | \pi, s^{0} \sim \phi) \nonumber \\ 
	& = P(s^{0} = s | \pi, s^{0} \sim \phi)  +  \nonumber \\
	& \qquad \qquad \qquad \sum_{t=1}^{\infty} \gamma^{t} P(s^{t} = s | \pi, s^{0} \sim \phi) \nonumber \\
	& = \phi(s)  + \sum_{t=0}^{\infty} \gamma^{t+1} P(s^{t+1} = s | \pi, s^{0} \sim \phi) \nonumber \\
	& = \phi(s)  + \gamma \sum_{t=0}^{\infty} \gamma^{t} P(s^{t+1} = s | \pi, s^{0} \sim \phi) \nonumber  \\
	& = \phi(s)  + \gamma \int_{S} \int_{A} \pi(a|s') P_a(s', s) \nonumber \\
	&\qquad \qquad \qquad \sum_{t=0}^{\infty} \gamma^{t} P(s^{t} = s' | \pi, s^{0} \sim \phi) dads' \nonumber \\
	& =  \phi(s) + \gamma \int_{S} \int_{A} \pi(a|s') P_a(s', s) \rho^{\pi}(s') dads' \nonumber
\end{align}	

\subsection{Proof of Lemma 1}
The Lagrangian of \eqref{eqn:density_constrained_dual} is
\begin{multline}
\mathcal{L}=\int_S\int_A r(s,a)\bar{\rho}_{s}^{\pi}(s,a)dads - \\ \int_S\int_A \mu(s,a)\Big(\bar{\rho}_{s}^{\pi}(s,a) - \pi(a|s) (\phi(s)+ \\ \gamma \int_S\int_A P_{a'}(s',s)\bar{\rho}_{s}^{\pi}(s',a')da'ds')\Big) dads
\end{multline}
where $\mu:S\times A\to \mathbb{R}$ is the Lagrange multiplier. The key step is by noting that
\small
\begin{multline}
\int_S\int_A\int_S\int_A \mu(s,a)\pi(a|s)P_{a'}(s',s)\bar{\rho}_{s}^{\pi}(s',a')da'ds'dads \equiv \\ \int_S\int_A\int_S\int_A \mu(s',a')\pi(a'|s')P_{a}(s,s')\bar{\rho}_{s}^{\pi}(s,a)da'ds'dads \nonumber
\end{multline}
\normalsize
Then by rearranging terms, the Lagrangian becomes 
\begin{multline}
\mathcal{L}=\int_S \phi(s) \int_A \mu(s,a)\pi(a|s) da ds- \\ \int_S\int_A \bar{\rho}_{s}^{\pi}(s,a)\Big(\mu(s,a)-r(s, a) - \\ \gamma \int_S P_{a}(s, s') \int_A \pi(a'|s') \mu(s',a') da'ds'\Big)dads 
\end{multline}
By the KKT condition and taking $Q=\mu$, the optimality condition satisfies \eqref{eqn:density_constraint} exactly.

\subsection{Proof of Theorem 1}
The solution $\pi$ to the primal problem is the optimal policy for the modified MDP with reward $r + \sigma_--\sigma_+$, which means $\pi$ is the optimal solution to:
\begin{align}
&\mathop{\max}\limits_{Q,\pi} \int_S \phi(s) \int_A Q^{\pi}(s,a)\pi(a|s) da ds  \nonumber \\
&\mathrm{s.t.}~ Q^{\pi}(s,a)= r(s, a) + \sigma_{-}(s) - \sigma_{+}(s) + \nonumber \\
& \qquad \qquad \qquad \gamma \int_S P_{a}(s, s') \int_A \pi(a'|s') Q^{\pi}(s',a') da'ds' \nonumber
\end{align}
$\pi$ is also the optimal solution to:
\begin{align}
&\mathop{\max}\limits_{\bar{\rho},\pi} \int_S\int_A\bar{\rho}^{\pi}(s,a) (r(s, a) + \sigma_{-}(s) - \sigma_{+}(s)) dads \nonumber \\
&\mathrm{s.t.}~\bar{\rho}^{\pi}(s,a)=\pi(a|s) \Big(\phi(s)+ \nonumber \\
& \qquad \qquad \qquad \gamma \int_S\int_A P_{a'}(s',s)\bar{\rho}^{\pi}(s',a')da'ds'\Big) \nonumber
\end{align}
Therefore, for any feasible policy $\pi'$ the following inequality holds:
\begin{multline}
	\int_S\int_A\bar{\rho}^{\pi}(s,a) (r(s, a) + \sigma_{-}(s) - \sigma_{+}(s)) dads \geq \\ \int_S\int_A\bar{\rho}^{\pi'}(s,a) (r(s, a) + \sigma_{-}(s) - \sigma_{+}(s)) dads \label{eqn:primal_optimal_inequality}
\end{multline}
By complementary slackness, if $\sigma_-(s)>0$, then $\rho^\pi(s)=\rho_{min}(s)$. The same applies to $\sigma_+$ and $\rho_{max}$. Since $\rho^{\pi'}(s) \geq \rho_{min}(s)$ and $\rho^{\pi'}(s) \leq \rho_{max}(s)$, we have:
\begin{multline}
	\int_S\int_A\bar{\rho}^{\pi}(s,a) (\sigma_{-}(s) - \sigma_{+}(s)) dads \leq \\ \int_S\int_A\bar{\rho}^{\pi'}(s,a) (\sigma_{-}(s) - \sigma_{+}(s)) dads \label{eqn:lagrange_inequality}
\end{multline}
Then we use \eqref{eqn:lagrange_inequality} to eliminate the $\sigma_{-}(s) - \sigma_{+}(s)$ in (\ref{eqn:primal_optimal_inequality}) and derive:
\begin{align}
	\int_S\int_A\bar{\rho}^{\pi}(s,a) r(s, a) dads \geq \int_S\int_A\bar{\rho}^{\pi'}(s,a) r(s, a) dads \nonumber
\end{align}
which means $\pi$ is the optimal solution maximizing $J_d^\star$ among all the solutions satisfying the density constraints. As a result, $\pi$ is the optimal solution to the DCRL problem.

\subsection{Proof of Lemma 2}
The proof of Lemma 2 follows from the proof of Theorem 2.2.7 in Rockafellar~\yrcite{rockafellar1970convex}. For simplicity, let $\hat{g} = \hat{g}(\sigma_{+}, \alpha)$. For all $\sigma_+' \in \mathcal{P}$, we have
\begin{equation}
    \begin{aligned}
     &d(\sigma_+')-\frac{\mu_d}{2}||\sigma_+'-\sigma_+^k||^2 \\
     &\le d(\sigma_+^k)+\langle \nabla d(\sigma_+^k), \sigma_+'-\sigma_+^k\rangle\\
     &= d(\sigma_+^k)+\langle \nabla d(\sigma_+^k),y^+-\sigma_+^k\rangle+ \\
     & \qquad \qquad \qquad \langle \nabla d(\sigma_+), \sigma_+'-\sigma_+^{k+1}\rangle\\
     &\le d(\sigma_+^k)+\langle \nabla d(\sigma_+^k),\sigma_+^{k+1} -\sigma_+^k\rangle +  \\
     & \qquad \qquad \qquad \langle \hat{g},\sigma_+'-\sigma_+^{k+1}\rangle + \sqrt{\frac{2\epsilon}{\mu}}||\sigma_+'-\sigma_+^{k+1}||\\
     &\le d(\sigma_+^{k+1})-\frac{\alpha}{2}||\hat{g}||^2+ \\
     & \qquad \qquad \qquad \langle\hat{g}, \sigma_+'-\sigma_+^{k} \rangle + \sqrt{\frac{2\epsilon}{\mu}}||\sigma_+'-\sigma_+^{k+1}|| \nonumber
    \end{aligned}
\end{equation}
Taking $\sigma_{+}'=\sigma_+^{k}$ and utilizing the fact that $\sigma_+^{k+1}-\sigma_+^{k}=\alpha \hat{g}$ gives the result.

\subsection{Proof of Theorem 2}
Let $\phi(\sigma_+) = \min_{\sigma_+'\in \mathcal{P}^{\star}} ||\sigma_+(s) - \sigma_+'||$. Based on the Theorem 4.1 of Luo and Tseng~\yrcite{luo1993convergence}, there exists a constant $\tau$ such that
\begin{equation}
    \phi(\sigma_+^{k})+||\rho^\star-\rho||\le \tau ||\sigma_+^{k+1}-\sigma_+^{k}||
\end{equation}
This shows that the distance to $\mathcal{P}^\star$ decreases linearly with $\hat{g}$.  From Lemma 2, it is clear that $d(\sigma_+)$ monotonically increases for a sequence generated by Algorithm 1 when $||\hat{g}||>2\frac{1}{\alpha}\sqrt{\frac{2\epsilon}{\mu}}$, and the imperfect dual ascent would reach a $\hat{\sigma}_+$ satisfying $||\hat{g}(\hat{\sigma}_+,\alpha)||<\frac{2}{\alpha} \sqrt{\frac{2\epsilon}{\mu}}$. From the fact that projection does not increase the distance between vectors, $||g(\sigma_{+}, \alpha)||<(\frac{2}{\alpha}+1) \sqrt{\frac{2\epsilon}{\mu}}$. Taking $\lambda = \sqrt{2} \tau (2 + \alpha)$, then it follows that $\min_{\sigma_+'\in \mathcal{P}^{\star}} ||\hat{\sigma}_+ - \sigma_+'|| \leq \lambda \sqrt{\frac{\epsilon}{\mu}}$. Then we have
\begin{equation}
    \begin{aligned}
    &d(\hat{\sigma}_+) = d(\sigma_+') + \int_0^1 \nabla d(\sigma_+' + t(\hat{\sigma}_+ - \sigma_+')) \mathrm{d}t(\hat{\sigma}_+ - \sigma_+')\\
    &\leq d(\sigma_+') + \int_0^1 \nabla d(\sigma_+') + \frac{1}{\mu} (\hat{\sigma}_+ - \sigma_+')t \mathrm{d}t(\hat{\sigma}_+ - \sigma_+') \\
    & = d(\sigma_+') +  \int_0^1 \nabla d(\sigma_+')(\hat{\sigma}_+ - \sigma_+')\mathrm{d}t + \frac{1}{2\mu} ||\hat{\sigma}_+ - \sigma_+'||^2 \\
    &= d(\sigma_+') + \frac{1}{2\mu} ||\hat{\sigma}_+ - \sigma_+'||^2 \nonumber
    \end{aligned}
\end{equation}
Also note that $d(\hat{\sigma}_+) - d(\sigma_+') \geq 0$. Thus we have:
\begin{equation}
    \begin{aligned}
    \min_{\sigma_+'\in \mathcal{P}^{\star}} ||d(\hat{\sigma}_+) - d(\sigma_+')|| \leq \lambda^2 \frac{1}{2\mu} \frac{\epsilon}{\mu} = \frac{\lambda^2}{2} \frac{\epsilon}{\mu^2
    }
    \end{aligned}
\end{equation}

\section{Supplementary Experiments}
In this section, we provide additional case studies that are not covered in the main paper. We mainly compare with RCPO~\citep{tessler2019reward} and the unconstrained DDPG~\citep{lillicrap2016ddpg}, which serves as the upper bound of the reward that can be achieved if the constraints are ignored.

\subsection{Express Delivery Service}
\begin{figure}[htbp]
	\centering
	\includegraphics[width=0.47\textwidth]{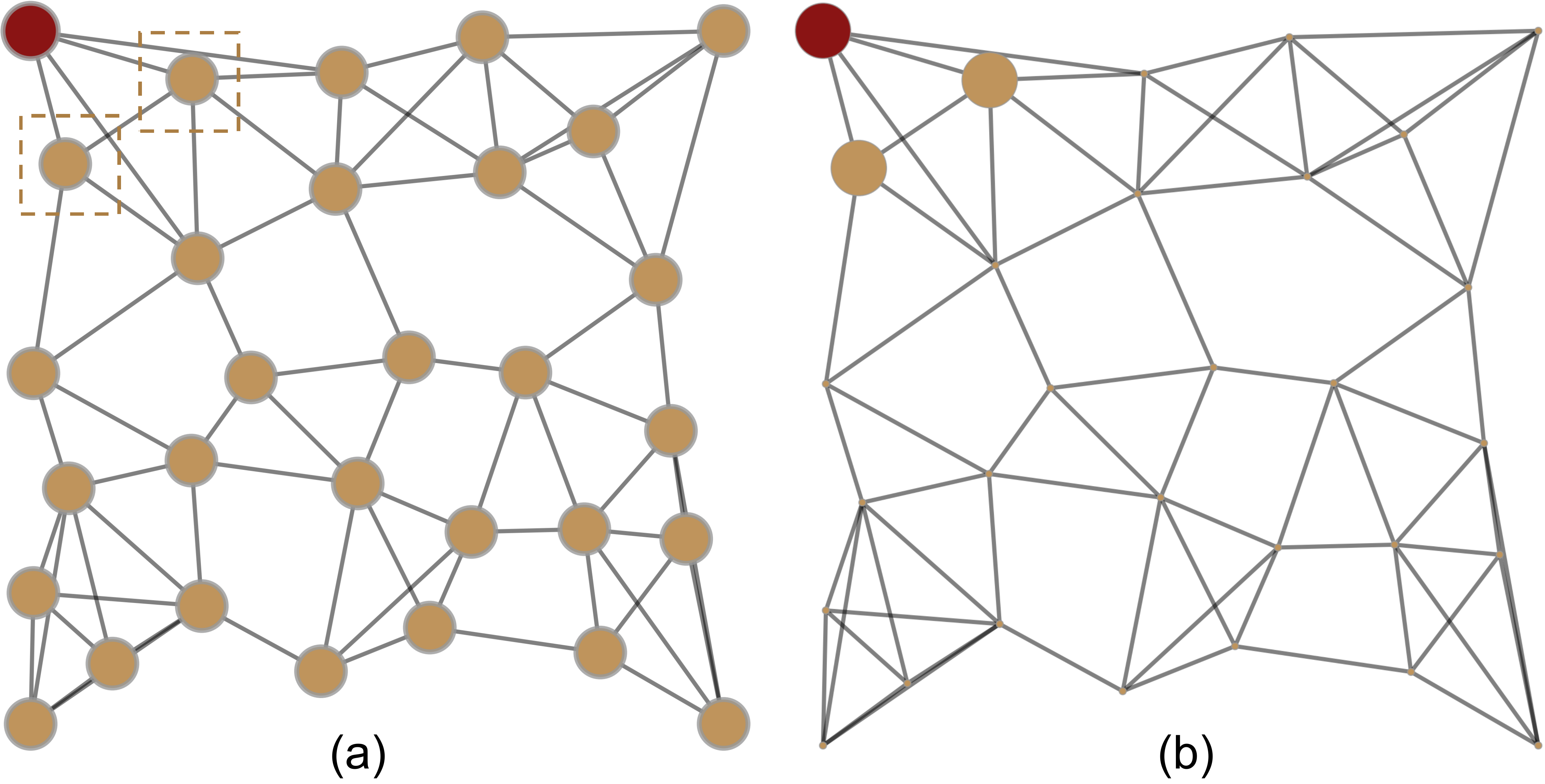}
	\caption{\footnotesize An example of the express delivery service company's transportation network with one ship center (red) and 29 service points (gold). (a) The vans start from the service points bounded by squares with equal probability, then visit other service points following a transition probability (policy), and finally reach the ship center (goal). (b) The standard Q-Learning method finds a policy that drives the vans directly to the goal without visiting any other service points, which minimizes the cost (traveling distance). The sizes of gold nodes represent the state density.}
	\label{fig:express_net}
\end{figure}

\begin{table*}[t]
	\centering
	\caption{\footnotesize Results of the express delivery service transportation task. The maximum allowed running time to solve for a feasible policy is 600s. The cost is the expectation of traveling distance from initial states to the goal.}
	\resizebox{\textwidth}{!}{
		\setlength{\tabcolsep}{3.5mm}{
			\begin{tabular}{c r r r r r r r r r r}
				\toprule
				\multirow{2}{*}{Density space} & \multirow{2}{*}{Method} &  \multicolumn{3}{c}{$\rho_{min} = 0.1$} & \multicolumn{3}{c}{$\rho_{min} = 0.3$} & \multicolumn{3}{c}{$\rho_{min} = 0.5$} \\ \cmidrule{3-11}
				&                         &  Solved  & Time (s) & Cost              &  Solved  & Time (s) & Cost             &  Solved  & Time (s) & Cost \\ \midrule
				\multirow{2}{*}{$\rho_s \in \mathbb{R}^{10}$}    & CERS  &  True    & 163.69   & 3.85              &  True    & 183.81   & 5.36             & True     & 466.32   & 5.12 \\
				& DCRL  &  True    & 1.35     & 2.91              &  True    & 2.05     & 2.66             & True     &  4.31    & 4.28  \\  \midrule
				\multirow{2}{*}{$\rho_s \in \mathbb{R}^{20}$}    & CERS  &  True    & 227.15   & 5.80              &  True    & 527.26   & 6.00             & False    & Timeout  & -     \\
				& DCRL  &  True    & 3.41     & 5.58              &  True    & 3.99     & 6.16             & True     & 5.28     & 6.27  \\ \midrule
				\multirow{2}{*}{$\rho_s \in \mathbb{R}^{100}$}   & CERS  &  True    & 161.62   & 10.26             & False    & Timeout  & -                & False    & Timeout  & -     \\
				& DCRL  &  True    & 3.53     & 10.23             & True     & 116.24   & 12.13            & True     & 153.86   & 14.06 \\
				\bottomrule
	\end{tabular}}}
	\label{tab:express}
\end{table*}

\begin{figure*}[t]
    \centering
    \includegraphics[width=0.95\textwidth]{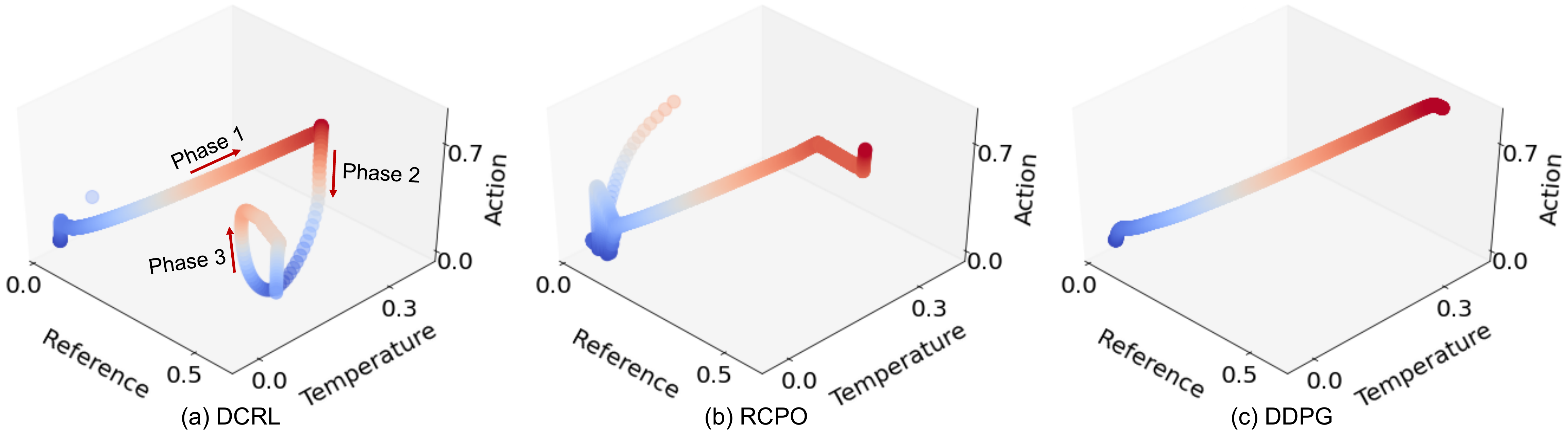}
    \caption{\footnotesize Visualization of the behavior of three methods in the safe electrical motor control task.}
    \label{fig:agent_behavior}
\end{figure*}

An express delivery service company has several service points and a ship center in a city. An example configuration is illustrated in \fig{\ref{fig:express_net}}~(a). The company uses vans to transport the packages from each service point to the ship center. The vans start from some service points following an initial distribution, travel through some service points and finally reach the ship center. The cost is formulated as the traveling distance. The frequency that each service point is visited by vans should exceed a given threshold in order to transport the packages in the service points to the ship center. Such frequency constraints can be naturally viewed as density constraints. A policy is represented as the transition probability of the vans from one point to surrounding points. The optimal policy should satisfy the density constraints and minimize the transportation distance.

This case study is proposed to further understand Algorithm~1 and its key steps. In Algorithm~1, our approach adds Lagrange multipliers to the original reward in order to compute a policy that satisfies density constraints. The update of Lagrange multipliers follows the dual ascent, which is key to satisfying the KKT conditions. In this experiment, we try to update the Lagrange multipliers using an alternative approach and see how the performance changes. We replace the dual ascent with the cross-entropy method, where a set of Lagrange multipliers $\Sigma = [\sigma_1, \sigma_2, \sigma_3, \cdots]$ are drawn from an initial distribution $\sigma \sim Z(\sigma)$ and utilized to adjust the reward respectively, after which a set of policies $[\pi_1, \pi_2, \pi_3, \cdots]$ are obtained following the same procedure in Algorithm~1. A small subset of $\Sigma$ whose $\pi$ has the least violation of the density constraints are chosen to compute a new distribution $Z(\sigma)$, which is utilized to sample a new $\Sigma$. The loop continues until we find a $\sigma$ whose $\pi$ completely satisfies the density constraints. We call this cross-entropy reward shaping (CERS). We experiment with 10D, 20D and 100D state spaces (corresponding to 10, 20 and 100 service points in the road network), whose density constraints lie in $\mathbb{R}^{10}$, $\mathbb{R}^{20}$ and $\mathbb{R}^{100}$ respectively. The density constraint vector $\rho_{min}: S \mapsto \mathbb{R}$ is set to identical values for each state (service point). For example, $\rho_{min} = 0.1$ indicates the minimum allowed density at each state is $0.1$. In Algorithm~1, we use Q-Learning to update the policy for both DCRL and CERS since the state and action space are discrete.

From \tab{\ref{tab:express}}, there are two important observations. First, our computational time of finding the policy is significantly less than that of CERS. When $\rho_s \in \mathbb{R}^{10}$ and $\rho_{min} = 0.1$, our approach is at least 100 times faster than CERS on the same machine. When $\rho_s \in \mathbb{R}^{100}$ and $\rho_{min} = 0.5$, CERS cannot solve the problem (no policy found can completely satisfy the constraints) in the maximum allowed time (600s), while our approach can solve the problem in 153.86s. Second, the cost reached by our method is generally lower than that of CERS, which means our method can find better solutions in most cases.

\subsection{Safe Electrical Motor Control (Section 5.3)}

To gain more insight on the behavior of our DCRL agent, we visualize the trajectories and actions (duty cycles) taken at different temperatures and reference angular velocities in \fig{\ref{fig:agent_behavior}}. In \fig{\ref{fig:agent_behavior}}~(a), The trajectory using DCRL can be divided into three phases. In Phase 1, as the reference angular velocity grows, the duty cycle also increases, so the motor temperature goes up. When the temperature is too high, the algorithm enters Phase 2 where it reduces the duty cycle to control the temperature, even though the reference angular velocity remains high. As the temperature goes down, the algorithm enters Phase 3 and increases the duty cycle again to drive the motor angular velocity closer to the reference. In \fig{\ref{fig:agent_behavior}}~(b), when the temperature is high, the RCPO algorithm will stop increasing the duty cycle but will not decrease it as Algorithm~1 does. So the temperature remains high and thus the density constraints are violated. In \fig{\ref{fig:agent_behavior}}~(c), the unconstrained DDPG algorithm continues to increase the duty cycle in spite of the high temperature.

\begin{figure*}[t]
	\centering
	\includegraphics[width=\textwidth]{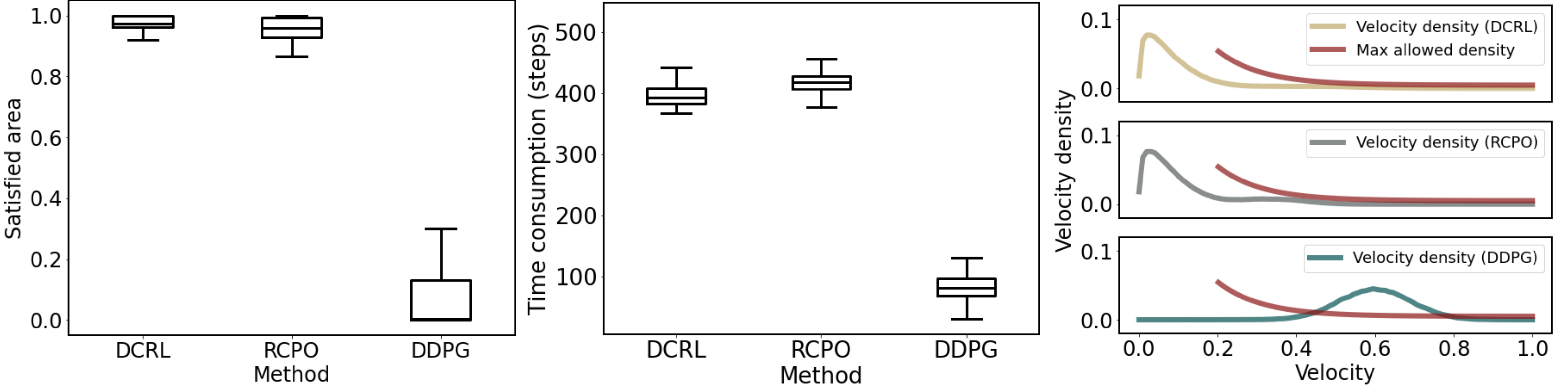}
	\caption{\footnotesize Results of the agricultural spraying problem with minimum pesticide density $(0, 1, 1, 0, 1)$ and maximum density $(0, 2, 2, 0, 2)$ from area $0$ to $4$. Left: Percentage of the entire area that satisfies the pesticide density requirement. Middle: Time consumption in steps. Whiskers in the left and middle plots denote confidence intervals. Right: visualization of the velocity densities using different methods.}
	\label{fig:farm_density_v1}
\end{figure*}
\begin{figure*}[t]
	\centering
	\includegraphics[width=\textwidth]{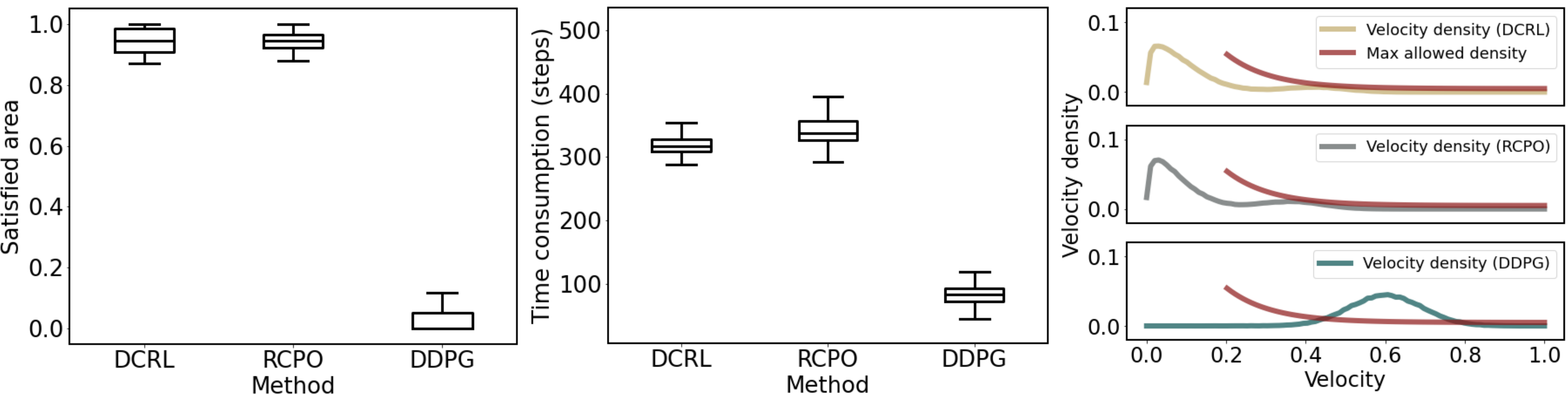}
	\caption{\footnotesize Results of the agricultural spraying problem with minimum pesticide density $(0, 0, 1, 1, 0)$ and maximum density $(0, 0, 2, 2, 0)$ from area $0$ to $4$. Left: Percentage of the entire area that satisfies the pesticide density requirement. Middle: Time consumption in steps. Whiskers in the left and middle plots denote confidence intervals. Right: visualization of the velocity densities using different methods.}
	\label{fig:farm_density_v2}
\end{figure*}

\subsection{Agricultural Spraying Drone (Section~5.4)}
In the agricultural pesticide spraying problem, we examine the methods with different pesticide density requirements and drone configurations to assess their capability of generalizing to new scenarios. In our main paper, from area 0 to 4, the minimum and maximum pesticide density are $(1, 0, 0, 1, 1)$ and $(2, 0, 0, 2, 2)$ respectively. In this supplementary material, we evaluate with two new configurations. In Figure~\ref{fig:farm_density_v1}, the minimum and maximum density are set to $(0, 1, 1, 0, 1)$ and $(0, 2, 2, 0, 2)$ from area $0$ to $4$. In Figure~\ref{fig:farm_density_v2}, the minimum and maximum density are set to $(0, 0, 1, 1, 0)$ and $(0, 0, 2, 2, 0)$ from from area $0$ to $4$. 

Although the settings are different from our main paper, the results convey consistent information. In Figure~\ref{fig:farm_density_v1}~and~\ref{fig:farm_density_v2}, DCRL and RCPO demonstrates similar performance in controlling pesticide densities to be within the minimum and maximum thresholds, while DCRL demands less time to finish the task. DDPG only minimizes the time consumption and thus requires the least time among the three methods, but cannot guarantee the pesticide density is satisfied. In terms of the velocity control, both DCRL and RCPO can avoid the high-speed movement. These observations suggest that when both DCRL and RCPO finds feasible policies satisfying density constraints, the policy found by DCRL can achieve lower cost or higher reward defined by the original unconstrained problem, which is the time consumption of executing the task in this case study.

\subsection{Mars Rover}

\begin{figure}[h]
    \centering
    \includegraphics[width=0.4\textwidth]{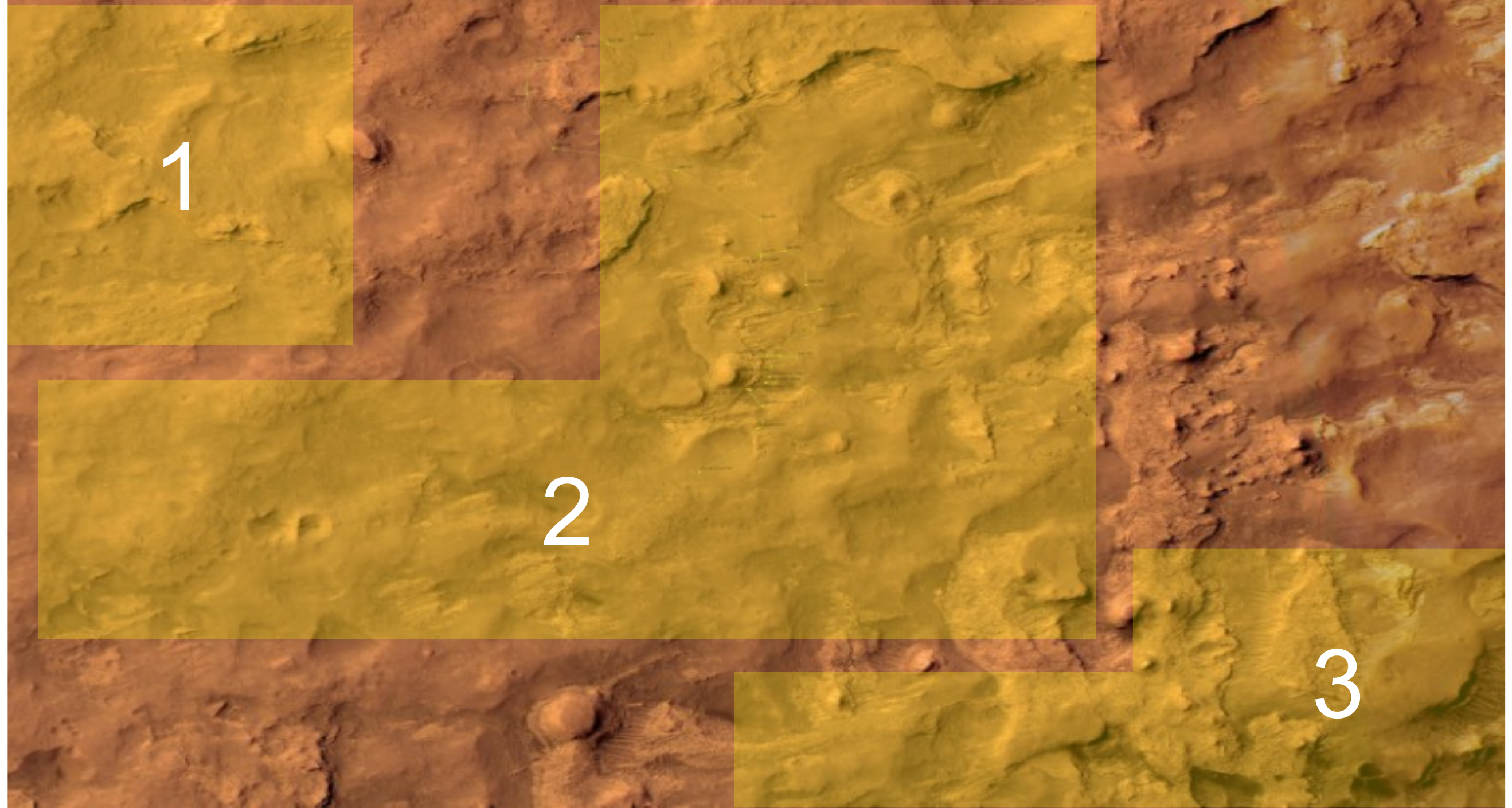}
    \caption{The mars rover environment. The agent starts from a random location in area 1 and is required to reach area 3. Area 2 is considered dangerous and the constraint is set on the total time that the agent is within area 2.}
    \label{fig:mars_rover}
\end{figure}

\begin{figure}[h]
    \centering
    \includegraphics[width=\linewidth]{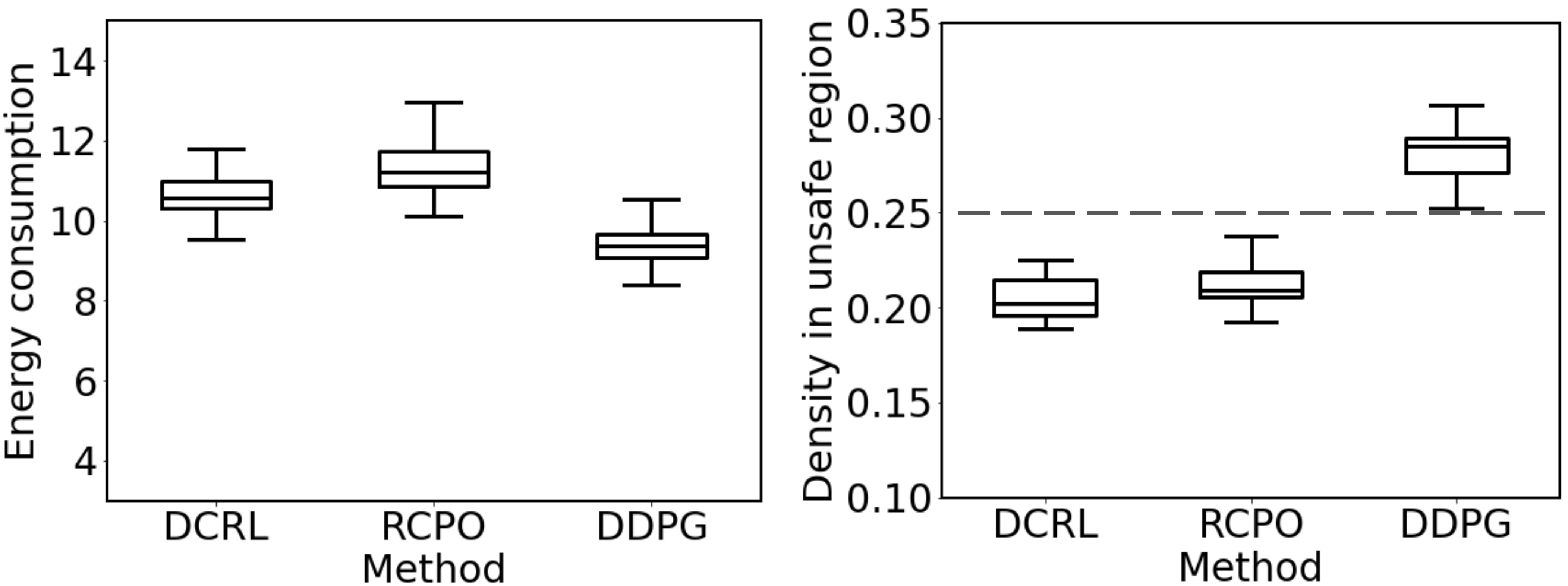}
    \caption{Results of the mars rover task.}
    \label{fig:mars_rover_result}
\end{figure}

We consider a mars rover task where the agent must constrain the amount of time within the dangerous region. The environment is shown in Figure~\ref{fig:mars_rover}, where there are three areas marked in yellow. The agent starts from a random location in area 1 and needs to reach area 3. At each timestep, the agent will receive a negative reward proportional to the energy consumption rate, and will receive a +10 reward after it reaches area 3. Area 2 is considered dangerous and the amount of time that the agent stays in area 2 is constrained by an upper bound. In the RCPO method, the agent receives a negative reward if it is inside area 2, and the magnitude of this negative reward is automatically tuned by RCPO itself. In our DCRL method, the constraint is converted to an equivalent state density constraint in area 2.

The objective is to minimize the energy consumption while respecting the time constraint in area 2. Figure~\ref{fig:mars_rover_result} shows the performance of the 3 methods. It is shown that DCRL and RCPO have similar energy consumption and both satisfy the density constraint. DDPG only minimizes the energy consumption and does not consider the constraint in area 2. 

\end{document}